\documentclass{article}

\PassOptionsToPackage{numbers}{natbib}

\usepackage[preprint]{neurips_2025}

\usepackage[utf8]{inputenc} 
\usepackage[T1]{fontenc}    
\usepackage{hyperref}       
\usepackage{url}            
\usepackage{booktabs}       
\usepackage{amsfonts}       
\usepackage{nicefrac}       
\usepackage{microtype}      
\usepackage{xcolor}         
\usepackage{multirow}						
\usepackage{makecell}
\usepackage{svg}
\usepackage{enumitem}
\usepackage{comment}

\title{GraphRAFT: Retrieval Augmented Fine-Tuning for Knowledge Graphs on Graph Databases}

\author{%
  Alfred Clemedtson \\
  Neo4j\\
  \texttt{alfred.clemedtson@neo4j.com} \\
  \And
  Borun Shi \\
  Neo4j \\
  \texttt{brian.shi@neo4j.com} \\
}

\usepackage{xcolor}

\usepackage{ifxetex}
\usepackage{ifluatex}
\newif\ifxetexorluatex 
\ifnum 0\ifxetex 1\fi\ifluatex 1\fi>0
   \xetexorluatextrue
\fi

\ifxetexorluatex
  \usepackage{fontspec}
\else
  \usepackage[T1]{fontenc}
  \usepackage[utf8]{inputenc}
\fi

\usepackage{textcomp}
\usepackage{xcolor}
\usepackage{listings}
\usepackage{upquote}

\definecolor{keyword}{HTML}{2771a3}
\definecolor{pattern}{HTML}{b53c2f}
\definecolor{string}{HTML}{be681c}
\definecolor{relation}{HTML}{7e4894}
\definecolor{variable}{HTML}{107762}
\definecolor{comment}{HTML}{8d9094}

\lstdefinelanguage{cypher}
{
	morekeywords={
		MATCH, OPTIONAL, WHERE, NOT, AND, OR, XOR, RETURN, DISTINCT, ORDER, BY, ASC, ASCENDING, DESC, DESCENDING, UNWIND, AS, UNION, WITH, ALL, CREATE, DELETE, DETACH, REMOVE, SET, MERGE, SET, SKIP, LIMIT, IN, CASE, WHEN, THEN, ELSE, END,
		INDEX, DROP, UNIQUE, CONSTRAINT, EXPLAIN, PROFILE, START,
            SELECT, FROM, JOIN, ON
	}
}

\newcommand{\mycdots}{\cdot\!\cdot\!\cdot}
\usepackage{graphicx}
\usepackage{amsmath}
\usepackage{amsthm}
\newtheorem{lemma}{Lemma}
\usepackage{subcaption}  
\usepackage{soul}

\begin{document}

\maketitle

\begin{abstract}
Large language models have shown remarkable language processing and reasoning ability but are prone to hallucinate when asked about private data. Retrieval-augmented generation (RAG) retrieves relevant data that fit into an LLM's context window and prompts the LLM for an answer. GraphRAG extends this approach to structured Knowledge Graphs (KGs) and questions regarding entities multiple hops away. The majority of recent GraphRAG methods either overlook the retrieval step or have ad hoc retrieval processes that are abstract or inefficient. This prevents them from being adopted when the KGs are stored in graph databases supporting graph query languages. In this work, we present GraphRAFT, a retrieve-and-reason framework that finetunes LLMs to generate provably correct Cypher queries to retrieve high-quality subgraph contexts and produce accurate answers. Our method is the first such solution that can be taken off-the-shelf and used on KGs stored in native graph DBs. Benchmarks suggest that our method is sample-efficient and scales with the availability of training data. Our method achieves significantly better results than all state-of-the-art methods across all four standard metrics on two set of challenging Q\&As on large text-attributed KGs.

\end{abstract}

\section{Introduction}

Large language models (LLMs) have shown remarkable ability to reason over natural languages. However, when prompted with questions regarding new or private knowledge, LLMs are prone to hallucinations\cite{huang2025survey}. A popular approach to remediate this issue is Retrieval-Augmented Generation (RAG) \cite{lewis2020retrieval}. Given a natural language question and a set of text documents, RAG retrieves a small set of relevant documents and feeds them into an LLM.

In addition to unstructured data such as text documents, many of the real-world structured data come in the form of Knowledge Graphs (KGs). KGs can be used to model a variety of domains such as the Web, social networks, and financial systems. Knowledge Base Question Answering (KBQA)\cite{lan2021survey} studies the methods used to answer questions about KGs. Most of the work studies multi-hop questions on the Web (open-domain)\cite{Yih2016TheVO, freebase, talmor-berant-2018-web}. 

Historically, there are two categories of approaches for KBQA. One approach is based on semantic parsing\cite{yih-etal-2015-semantic, luo-etal-2018-knowledge, 10.1007/978-3-319-93417-4_46, 10.1145/2806416.2806472, chen-etal-2019-uhop, 10.1145/3357384.3358033, lan-jiang-2020-query}, which involves converting natural language questions into logical forms that are often SPARQL\cite{sparql}. Another approach is based on information retrieval\cite{sun2019pullnetopendomainquestion, sun-etal-2018-open, xiong-etal-2019-improving, he-etal-2017-generating}, which involves embedding and retrieving triplets from KGs and reranking the extracted data.

There is a growing demand to leverage the language understanding and reasoning power of LLMs to improve open-domain KBQA. At the same time, GraphRAG extends the RAG setup to private KGs. The methodologies began to merge, despite the different application areas. A unifying architecture follows the general framework of 1) identifying some entities in the given question, 2) retrieving a subgraph and 3) prompting an LLM for reasoning. 

\cite{baek2023knowledgeaugmentedlanguagemodelprompting, hu2024graggraphretrievalaugmentedgeneration} retrieves and reranks the top nodes (or k-hop ego-subgraphs) and prompts the LLM. \cite{luo2024reasoning} prompts an LLM to generate relation path templates and retrieves concrete paths using in-memory breadth-first search and fintunes the LLM. \cite{sun2024thinkongraph, ma2024think, sui2024fidelisfaithfulreasoninglarge, wang2025reasoning, xia2024knowledgeawarequeryexpansionlarge} identify starting entities and iteratively perform the retrieval step-by-step by prompting the LLM what is the next step to take. \cite{he2024gretriever} prunes a given subgraph with parameterised graph algorithms and prompts an LLM with a textualised version of the local subgraph. \cite{wang2023knowledgedrivencotexploringfaithful} extends LLM chain-of-thought (CoT) by iteratively retrieving from the KG. \cite{chen2025knowledge} enhances the quality of the QA training data.
 
Although significant progress has been made, there are several limitations of existing methods. Many of the real-world KGs are stored in native graph DBs. Graph DBs come with query engines that optimise user queries, such as Cypher. However, none of the existing methods leverage such graph DBs at all. Any method that explicitly requires step-by-step retrieval (due to iteratively prompting the LLMs) prevents direct multi-hop querying on DBs. This implies that each step of the retrieval process needs to be fully materialised and the fixed sequential order of traversal implementation is frequently suboptimal. Benchmarking is often performed with in-memory graphs represented as tensors with adhoc implementations of subgraph retrieval.

On the other hand, many methods that involve multi-hop retrieval as a single step remain abstract. They usually stop at the point of specifying some path patterns. Converting the abstract path definitions into queries remains a separate challenging task, irrespective of specific query languages\cite{lei2025spider20evaluatinglanguage, texttosparql, brei2024leveragingsmalllanguagemodels, ozsoy2024text2cypherbridgingnaturallanguage}. Another stream of work explicitly requires the questions to be of some logic fragment and converts them into SPARQL. There is also no general way of guaranteeing that the generated queries are syntactically and semantically correct with respect to specific KGs.

In this paper, we propose a simple, modular, and reliable approach for question answering on private KGs (GraphRAG). Specifically, the contributions of our papers are:
\begin{itemize}
    \item We propose a new method of fine-tuning an LLM to generate optimal Cypher queries that only requires textual QAs as training data but not user-crafted ground-truth queries.
    \item At inference time, we deploy a novel grounded constrained decoding strategy to generate provably syntactically and semantically correct Cypher queries.
    \item Our method follows a modular and extensible retrieve-and-reason framework and can be taken off-the-shelf to perform GraphRAG on native Graph DBs.
\end{itemize}
In addition, benchmarks show that our method achieves significantly beyond SOTA results across all 4 metrics on STaRK-prime and STaRK-mag, two large text-attributed KGs with Q\&As for GraphRAG. For example, our method achieves 63.71\% Hit@1 and 68.99\% Mean Reciprocal Rank (MRR), which is an improvement of 22.81 and 17.79 percentage points, respectively, over the best previous baselines on STaRK-prime.
Our method is sample-efficient, achieving beyond SOTA results on all metrics on STaRK-prime when trained with only 10\% of the data, and scales with more training data.

\section{Related Work}

\textbf{RAG.} Recent work extends RAG\cite{lewis2020retrieval} to broader settings. RAFT\cite{raft2024} studies domain-specific retrieval-augmented fine-tuning on by sampling relevant and irrelavant documents. GraphRAG\cite{edge2024local} generalises RAG to global text summarisation tasks in multiple documents. The term GraphRAG has also been widely adopted to mean RAG on (knowledge) graphs, which is the problem setup we study in this paper. \cite{graft} applies RAFT for GraphRAG on document retrieval and summarisation. There are many methodological and implementation improvements\cite{li2024simple} and industrial applications of GraphRAG\cite{Xu_2024}.

\textbf{Message Passing Graph Neural Networks and Graph Transformers.} Message-passing Graph Neural Networks (GNNs) iteratively aggregates and updates embeddings on nodes and edges using the underlying graph as the computation graph, hence incorporating strong inductive bias. GNNs have been used within the framework of GraphRAG to improve retrieval\cite{mavromatis2024gnn} or to enhance the graph processing power of LLMs\cite{zhang2022greaselm, yasunaga2021qa}. Alternative model architectures such as Graph Transformers, which employ graph-specific tokenization or positional embedding techniques, have been developed\cite{ying2021transformersreallyperformbad, tokengt}. There are also work\cite{zhao2023learning, he2024harnessing, jiang2024ragraph} that leverage LLMs to improve classical graph problems such as node classification and link prediction on text-attributed graphs.

\section{Preliminary}

\subsection{Graph database and graph query language}
Graph DB treats both nodes and edges as primary citizens of the data model. A common graph data model is Labeled Property Graph (LPG). LPGs support types and properties on nodes and edges in a flexible way. For example, nodes of the same type are allowed to have different properties, and nodes and edges can share the same types. LPGs can be used to model virtually all real-world KGs.

A Graph DB stores LPGs efficiently on disks. It also comes with a query engine that allows one to query the graph. A widely used query language is Cypher\cite{cypher}\cite{cyphersemantics} (and a variant openCypher\cite{opencypher}), which are implementations of the recent GQL\cite{gql}\cite{gqlpaper} standard. We will refer to the two interchangeably throughout this paper. The key ingredient of Cypher is graph pattern matching. A user can query the graph by matching on patterns (e.g paths of a certain length) and predicates (e.g filtering on node and edge properties). An example Cypher query is provided in Figure~\ref{fig:cypher-example}. The execution of a query is carried out by the query engine which heavily optimises the order of executing low-level operators.

An alternative graph data model is Resource Description Framework (RDF)\cite{rdf}. It was originally created to model the Web. A graph is modeled as a collection of triples, commonly referred to as subject-predicate-object, where each element can be identified with a URI (mimicing the Web). Query languages on top of RDFs include SPARQL\cite{sparql}. RDFs also support ontology languages such as Web Ontology Language (OWL)\cite{owl} to model and derive facts through formal logic. Much of the KBQA literature has taken inspiration from RDFs by defining a graph as a collection of triples and reasoning over it in a formal style. Widely used benchmarks such as WebQSP\cite{Yih2016TheVO} are exactly questions about the Web that are answerable by SPARQL.

Many open-source and commercial graph DBs support Cypher over LPGs, such as Neo4j\cite{neo4j}, ArangoDB\cite{arangoDB}, TigerGraph\cite{tigergraph}. There are also popular RDF stores such as Neptune\cite{neptune}, which also supports Cypher over RDFs. 

\begin{figure}
 \centering
 \begin{lstlisting}[language=cypher]
    UNWIND $source_names AS source_name
    MATCH (src {name: source_name})-[r1]-(var)-[r2]-(tgt) WHERE src <> tgt
    RETURN labels(src)[1] AS label1, src.name AS name1, type(r1) AS type1, 
            labels(var)[1] AS label2, type(r2) AS type2, labels(tgt)[1] AS label3, 
            size([t IN collect(DISTINCT tgt) WHERE t.nodeId in $target_ids | t]) AS 
            correctCnt, count(DISTINCT tgt) AS totalCnt
 \end{lstlisting}
 \caption{An example Cypher query. It takes as user input a list variable source\_names and another list target\_ids. It iterates through them and find all two-hop neighbours of each source\_name node, requiring the second-hop node to be distinct to the source. The query returns aggregate information of the subgraph such as labels and types of nodes and edges, and arithmetic over how many second-hop nodes have ids that are in the user-defined node id list.}
 \label{fig:cypher-example}
\end{figure}

\subsection{LLM}
Modern LLMs are usually auto-regressive decoder-only Transformers as backbones that are trained on the Web. An LLM $f_\theta$ has a fixed vocabulary set $\mathcal{V}$. Given a sequence of characters $c_0, \cdots, c_n$, a tokenizer converts it to a sequence of tokens $t_0, \cdots, t_k$ where $t_i \in \mathbb{R}^d$ and $d$ is the embedding dimension. The tokenizer often compresses multiple characters into one token and splits a word into multiple tokens. Given $t_0, \cdots, t_k$, suppose $f_\theta$ has generated tokens $t_{k+1}, \cdots, t_{k+n}$, it computes the logits for the $k+n+1^{\text{th}}$ token as $l_{k+n+1} = f_\theta(t_0;t_{k+n})$ where $l_{k+n+1} \in \mathbb{R}^{|\mathcal{V}|}$. The probability of generating any token $x$ is obtained by applying softmax to the logits, $p(t_{k+n+1} = x | t_0;t_{k+n}) = exp(l_{k+n+1}^x) / \sum_{y \in \mathcal{V}}{exp(l_{k+n+1}^y)}$. Greedy decoding picks the token with the highest probability at each step. Beam search with width $m$ keeps $m$ sequences of tokens with the highest product of probabilities so far. The generative process normally terminates when some end-of-sequence $<eos>$ token is decoded. Pretrained LLMs can be finetuned efficiently using techniques such as LoRA\cite{hu2022lora}, optimising the product of conditional probabilities of next-token prediction.

\section{Approach} \label{sec:approach}
Let $\mathcal{G} = (V, E, L, l_v, l_e, K, W, p_v, p_e)$ be a KG stored in a graph DB. $V, E$ are nodes and edges. $L$ is a set of labels. $l_v: V \rightarrow \mathcal{P}(L)$ the label assignment function for $V$. $K, W$ the set of property keys and values. $p_v: V \rightarrow K \times W$ the property key-value assignment function on the nodes. $l_e, p_e$ are defined analogously for edges. Note that we support the most flexible definition of KGs, where nodes and edges can share labels, and those of the same label can have different properties keys. We assume that each node has at least one property that is text, which is common in the GraphRAG setup. All nodes $v$ are equipped with an additional property which we abbreviate as $z_{v}$, which is the text embedding produced by some text embedder $LM$ on the text attribute.

Given a set of training QAs $\{Q_i, A_i\}$ where $A_i \subset V$, we want our model to produce good answers $A_j$ for unseen questions ${Q_j}$ according to a variety of metrics. Our approach consists of several steps. First, we create a set of training question-Cypher pairs $\{Q_i, C_i\}$ to finetune a LLM (Section~\ref{sec:synthesize}). At inference time, we deploy a simple method of constrained decoding that is grounded to $\mathcal{G}$ to guarantee syntactically and semantically optimal Cypher queries (Section~\ref{sec:cd}). The optimal Cypher is executed to retrieve a text-attributed subgraph for each question and a second LLM is finetuned to jointly reason over text and subgraph topology to select the final answers (Section~\ref{sec:llm2}).

\subsection{Synthesize ground-truth cypher} \label{sec:synthesize}
For a given $Q_i$, we few-shot prompt an LLM $L_0$ to identify the relevant entities (strings) $n_1, \cdots n_k$ mentioned in the question. In order to address the inherent linguistic ambiguity and noise present in the questions and the graph, if a generated node name does not correspond to an existing node, we perform an efficient vector similarity search directly using vector index at the database level to retrieve the most similar nodes.
\begin{equation} \label{eq:knn}
v_i = cossim(LM(n_i), \{z_v\}_{v \in V})
\end{equation}

This enables more accurate identification of entities than performing native vector similarity between all nodes in the graph and an embedded vector representing the question.

We then construct several Cypher templates that match multi-hop subgraphs around the identified entities $v_1, \cdots, v_k$ with filters on node and relationship types. For each Cypher query $C_j$ we execute the query and compute the hits in the answers ${A_i}$, as well as the number of total nodes. The calculation (as an aggregation step) is performed as part of the query itself which means the nodes and edges and their properties do not need to be materialised and retrieved and hence saves memory and I/O workload. An example such query is shown in Figure~\ref{fig:cypher-example}.

\begin{figure}[h]
    \centering
    \includegraphics[width=\linewidth]{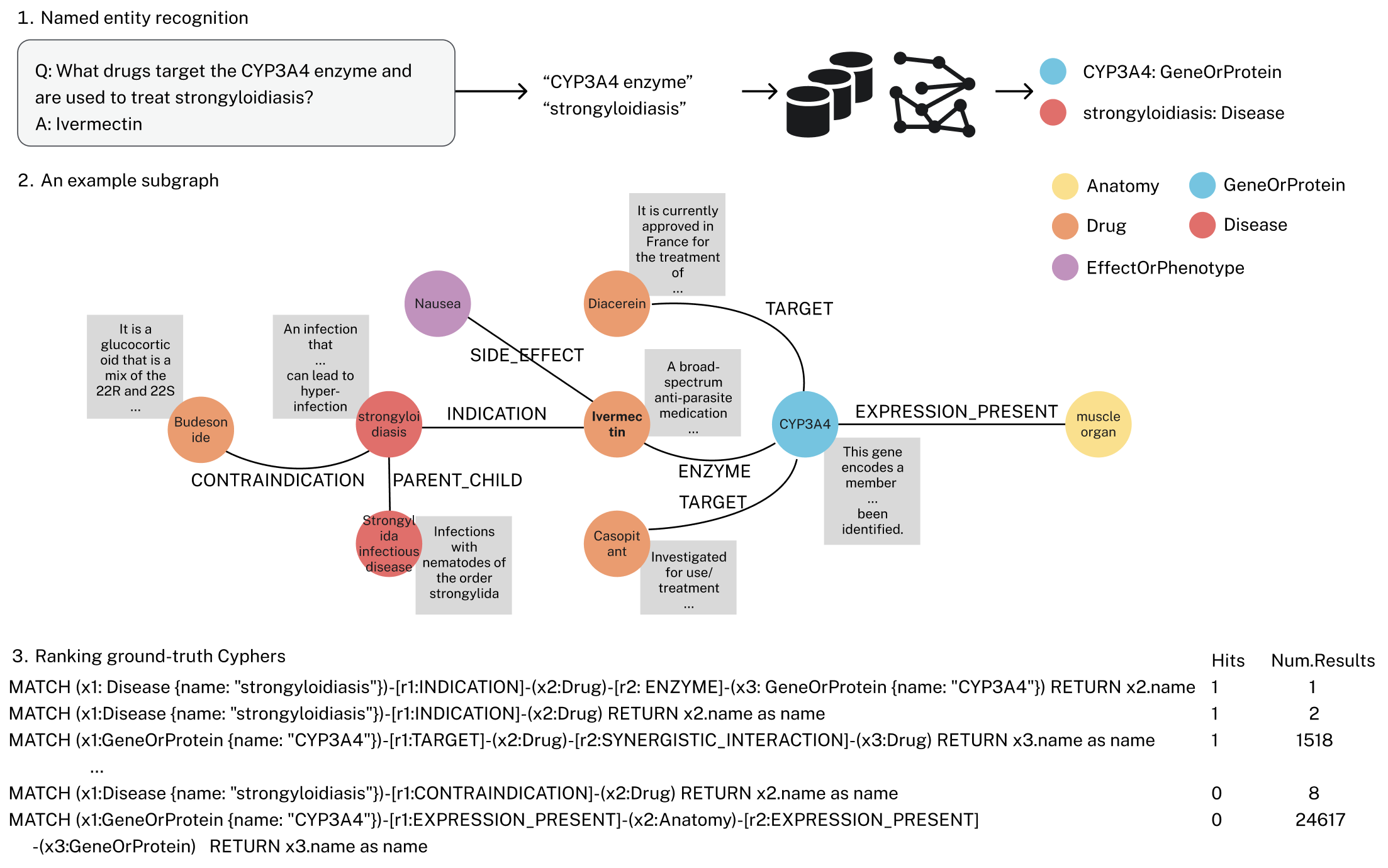}
    \caption{An example of creating ground-truth Cypher for a QA. In Step 1, few-shot LLM produces candidate entities which we ground with $\mathcal{G}$ in the DB with vector index. Step 2 shows part of the subgraph around the entity and answer nodes. With the DB, we execute the all one-hop, two-hop around each entity, and all length-two paths connecting the two entities in Step 3. We aggregate the hits and number of nodes for each query and rank them.}
    \label{fig:ner-cypher}
\end{figure}

We then rank the queries according to their hits and number of results and retain the top query $C_i$. This gives us a set of $\{Q_i,C_i\}$ between questions and best Cypher queries that contain the answers $\{A_i\}$. The best Cypher can be decided by reranking and for example filtering by recall and precision. We finetune an LLM $L_1$ with the given set of training data. The workflow is illustrated in Figure~\ref{fig:ner-cypher}.

\subsection{Grounded constrained decoding} \label{sec:cd}
Our $L_1$ has learned to generate graph pattern matching Cypher queries. However, there is still no guarantee that the generated Cypher is executable at inference time. The major bottlenecks are 1) syntactical correctness and 2) semantic correctness (i.e any type filters must correspond to existing node labels or edge types of $\mathcal{G}$). A common approach of generate-and-validate is both costly and slow and requires a sophisticated post-validation correction scheme to guarantee eventual correctness. 

We use a new method of next-token constrained decoding at the level of logits processor at inference time. The constraints are applied at the token-level instead of word (or query keyword) level since the tokenizers used by the majorioty of modern LLMs do not have a one-to-one mapping from words and query keywords to the tokens.

Given a question at inference time, we first create a set of possible m-hop queries involving identified entities. This step is efficient since the schema of the graph is available. Let there be $\mathcal{Q} = \{\mathcal{Q}^1, \cdots, \mathcal{Q}^M\}$ valid queries, each has tokenization $\mathcal{Q}^k = (n_0, \cdots, n_{Q_k}) \in \mathcal{V}^{Q_k}$ of variable length.
When $L_1$ has generated $i$ tokens $(t_0, \cdots, t_i)$ and is generating a vector of logits $(l_0, \cdots, l_\mathcal{V})$, our logits processor masks all invalid token with the value $-\infty$ by comparing with the $i+1^{\text{th}}$ token of all possible tokenized queries that match the initial $i$ tokens. The masking is defined as Equation~\ref{eq:masking} where $M_{i+1}$ is the set of valid tokens at $i+1^{\text{th}}$ position grounded in $\mathcal{G}$ via $\mathcal{Q}$.

\begin{equation} \label{eq:masking}
\tilde{\ell}_{i+1}^{(x)} = 
\begin{cases}
\ell_{i+1}^{(x)}, & \text{if } x \in M_{i+1} \\
-\infty, & \text{if } x \notin M_{i+1}
\end{cases}
\space \text{where }
M_{i+1} = \{t \in \mathcal{V} | \exists \mathcal{Q}^k \in \mathcal{Q} 
\text{ , } 
\mathcal{Q}^k_{0:i+1} = (t_0, \cdots, t_k, t) \}
\end{equation}

\begin{figure}[h]
    \centering
    \includegraphics[width=\linewidth]{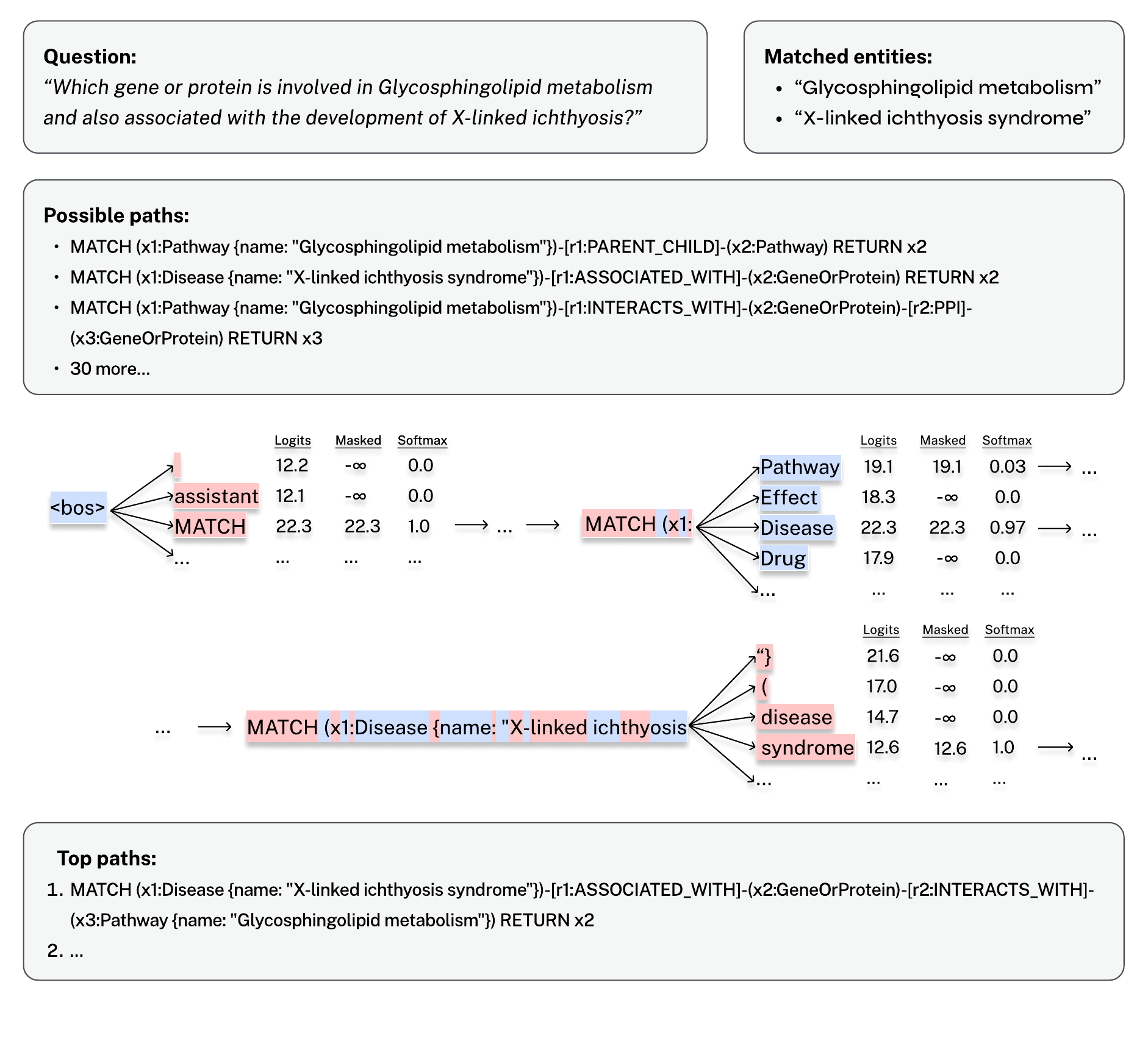}
    \caption{An example of grounded constrained decoding. For the given question, we tokenize all possible queries around it's identified entities. At each step during generation, our logits processor masks out invalid tokens. For example, after \textit{"ichthyosis"}, the LLM would have generated the symbols \textit{")} which has the highest logit. Our processor masks it out since this predicate \textit{{name: "X-linked ichthyosis"}} is invalid.}
    \label{fig:constrained-decoding}
\end{figure}

The tokenizer for $L_1$ then applies decoding after softmax. By construction, our decoded query is executable. This constraint decoding is faithful and non-invasive for both greedy decoding and beam search with sufficiently large beams, as stated in Lemma~\ref{lemma:cd}. The formal statement and proof are provided in Appendix~\ref{ap:cd}. Unlike existing constrained generation methods\cite{beurerkellner2024guidingllmsrightway, geng2024grammarconstraineddecodingstructurednlp}, our approach does not require any formal grammar to be defined and is context-aware with respect to $\mathcal{G}$.
An example of grounded constrained decoding is illustrated in Figure~\ref{fig:constrained-decoding}. Since our method is applied entirely on the logits before softmax, different final decoding such as greedy and beam search can be applied.

\begin{lemma} \label{lemma:cd}
(Informal) If a query is invalid, it will not be generated. When using beam width = 1, constrained decoding acts as greedy decoding among valid queries. When beam width = M, exactly all valid queries are generated.
\end{lemma}


\subsection{Finetuning LLM as local subgraph reasoner}\label{sec:llm2}
Given any question for KG $\mathcal{G}$, our trained LLM $L_1$ produces guaranteed executable and grounded Cypher queries. If we perform beam search with width $m$, we are able to obtain $m$ valid queries with highest total probabilities. For the simplest questions and sparse $\mathcal{G}$, the subgraph produced by the best query may be exactly the answer nodes. However, for more difficult multi-hop questions on dense $\mathcal{G}$, the subgraph still contains other relevant nodes (and edges) that are not the answers. In order to obtain only the answer nodes by reasoning over the small subgraph, which often requires reasoning on the textual properties beyond graph patterns, we train an LLM $L_2$ to perform the task. We construct the subgraph by executing $m$ queries until some threshold on the size of the graph or tokenization required to encode the graph. The output of $L_2$ can be viewed as selecting or reranking the nodes in the input textualised graph. An example prompt we use is provided in Figure~\ref{fig:prompt-example}.


\begin{figure}
\centering
\small
\begin{verbatim}
<|start_header_id|>user<|end_header_id|>
Given the information below, return the correct nodes for the following question: 
What drugs target the CYP3A4 enzyme and are used to treat strongyloidiasis?

Retrieved information:
pattern: ['(x1:GeneOrProtein {name: "CYP3A4"})-[r1:ENZYME]-
(x2:Drug {name: "Ivermectin"})-[r2:INDICATION]-(x3:Disease {name: "strongyloidiasis"})', 
'(x1:Disease {name: "strongyloidiasis"})-[r1:INDICATION]-(x2:Drug {name: "Ivermectin"})', 
'(x1:GeneOrProtein {name: "CYP3A4"})-[r1:ENZYME]-(x2:Drug {name: "Ivermectin"})']
name: Ivermectin
details: {'description': 'Ivermectin is a broad-spectrum anti-parasite medication. 
It was first marketed under the name Stromectol® and used against worms (except tapeworms), 
but, in 2012, it was approved for the topical treatment of head lice 
infestations in patients 6 months of age and older, 
and marketed under the name Sklice™ as well. ...

pattern: ['(x1:Disease {name: "strongyloidiasis"})-[r1:INDICATION]-
(x2:Drug {name: "Thiabendazole"})']
name: Thiabendazole
details: {'description': '2-Substituted benzimidazole first introduced in 1962. 
It is active against a variety of nematodes and is the drug of choice for 
strongyloidiasis. It has CNS side effects and hepatototoxic potential.  ...
...
<|start_header_id|>model<|end_header_id|>
\end{verbatim}
\caption{An example prompt that describes a local subgraph retrieved by Cypher queries around identified entities. This prompt contains both textual information and patterns used to retrieved them, which encodes topology information.}
\label{fig:prompt-example}
\end{figure}

\section{Experiments} \label{sec:exp}
\paragraph{Setup}
We use Neo4j as the graph database. The default database configuration is used. For the main result, we using OpenAI text-embedding-ada-002\cite{openai} as the text embedder $LM$ in Equation~\ref{eq:knn}, OpenAI gpt-4o-mini as LLM $L_0$ for few-shot entity resolution., gemma2-9b-text2cypher\cite{ozsoy2024text2cypher, team2024gemma} as our LLM $L_1$, Llama-3.1-8B-Instruct\cite{grattafiori2024llama} as our LLM $L_2$. All experiments are run on a single 40GB A100 GPU. Additional detailed experiment setup is provided in Appendix~\ref{sec:experiemnt_setup}. All of our code is available on Github\footnote{\url{https://github.com/AlfredClemedtson/graphraft}}.

\paragraph{Datasets}
We benchmark our method on the STaRK-prime and STaRK-mag datasets\cite{wu2025stark} (license: CC-BY-4.0). stark-prime is a set of Q\&As over a large biomedical knowledge graph PrimeKG\cite{chandak2023building}. The questions mimic roles such as doctors, medical scientists and patients. It contains 10 node types and 18 edge types with rich textual properties on the nodes. With 129k nodes and 8 million edges and at the same time a high density (average node degree 125), it serves as a challenging and highly suitable dataset to benchmark retrieval and reasoning on large real-world KGs.
stark-mag is a set of Q\&As over ogbn-mag\cite{hu2020open} that models a large academic citation network with relations between papers, authors, subject areas and institutions.

There is a large collection of Q\&A datasets on graphs, we benchmark on datasets most suitable to the GraphRAG domain\cite{bechler2025position} and elaborate on why we don't benchmark on the others. WebQ\cite{Berant2013SemanticPO}, WebQSP\cite{Yih2016TheVO}, CWQ\cite{talmor-berant-2018-web} ad GrailQA\cite{gu2021beyond} are popular KBQA benchmarks containing SPARQL-answerable few-hop questions over Freebase\cite{freebase}, a database of general knowledge with URIs on nodes. Our problem setting has no requirement on the form of the questions and targets private KGs instead of such largely wikipedia-based KG which LLMs are explicitly trained on. Freebase has been deprecated since 2015. HotpotQA\cite{yang2018hotpotqa} and BeerQA\cite{qi2021answering} are few-hop questions over Wikidata\cite{wikidata}. STaRK-amazon\cite{wu2025stark} models properties of products (such as color) as nodes (with has-color relation) and the product co-purchasing graph itself is homogeneous.

\subsection{Main results} \label{sec:main-results}
We show our result in Table~\ref{tab:main-result}. The baselines range from pure vector-based retrievers to GraphRAG solutions to agentic methods. We use the four metrics originally proposed in \cite{wu2025stark}. Hit@1 measure the ability of exact answering a right answer while the other three metrics provides more holistic view on the answer quality.

As is shown in Table~\ref{tab:main-result}, GraphRAFT gives best results on all metrics on both STaRK-prime and STaRK-mag. Even without using $L_2$ to reason over the local subgraph, our $L_1$ when used to retrieve nodes using generated Cypher queries (up to 20 nodes, for measuring recall), already gives better metrics than all SOTA methods on STaRK-prime.

\begin{table}
	\centering
	\caption{Main table of our results against previous baselines. Bold and underline represent the best method, underline represents the second best.}
	\label{tab:main-result}
	\begin{tabular}{c|cccc|cccc}
		\toprule
            & \multicolumn{4}{c|}{STARK-PRIME} & \multicolumn{4}{c}{STARK-MAG} \\
		  & Hit@1 & Hit@5 & R@20 & MRR & Hit@1 & Hit@5 & R@20 & MRR \\
		\midrule
            BM25\cite{Robertson2009ThePR} & 12.75 & 27.92 & 31.25 & 19.84 & 25.85 & 45.25 & 45.69 & 34.91 \\ 
            voyage-l2-instruct\cite{voyageai} & 10.85 & 30.23 & 37.83 & 19.99 & 30.06 & 50.58 & 50.49 & 39.66 \\
            GritLM-7b\cite{muennighoff2024generative} & 15.57 & 33.42 & 39.09 & 24.11 & 37.90 & 56.74 & 46.40 & 47.25 \\
            multi-ada-002\cite{openai} & 15.10 & 33.56 & 38.05 & 23.49 & 25.92 & 50.43 & 50.80 & 36.94 \\
            ColBERTv2\cite{santhanam-etal-2022-colbertv2} & 11.75 & 23.85 & 25.04 & 17.39 & 31.18 & 46.42 & 43.94 & 38.39 \\
            Claude3 Reranker (10\%) & 17.79 & 36.90 & 35.57 & 26.27 & 36.54 & 53.17 & 48.36 & 44.15 \\
            GPT4 Reranker (10 \%) & 18.28 & 37.28 & 34.05 & 26.55 & 40.90 & 58.18 & 48.60 & 49.00 \\
            HybGRAG\cite{lee2024hybgraghybridretrievalaugmentedgeneration} & 28.56 & 41.38  & 43.58 & 34.49 & \underline{65.40} & 75.31 & 65.70 & 69.80   \\
            AvaTaR\cite{wu2025avatar} & 18.44 & 36.73 & 39.31 & 26.73 & 44.36 & 59.66 & 50.63 & 51.15 \\
            MoR\cite{lei2025mixturestructuralandtextualretrievaltextrich} & 36.41 & 60.01 & 63.48 & 46.92 & 58.19 & 78.34 & 75.01 & 67.14 \\
            KAR\cite{xia2025knowledgeawarequeryexpansionlarge} & 30.35 & 49.30 & 50.81 & 39.22 & 50.47 & 65.37 & 60.28 & 57.51 \\
            MFAR\cite{li2024multifieldadaptiveretrieval} & 40.9 & 62.8 & 68.3 & 51.2 & 49.00 & 69.60 & 71.79 & 58.20  \\
            \midrule
            GraphRAFT w/o $L_2$ & 
            \underline{52.12} & \underline{71.55} & \underline{75.52} & \underline{60.72} &
            62.63 & 
            \textbf{\underline{86.68}} & 
            \underline{88.88} & 
            \underline{73.03} \\

            GraphRAFT & 
            \textbf{\underline{63.71}} & \textbf{\underline{75.39}} & \textbf{\underline{76.39}} & \textbf{\underline{68.99}} & 
            
            \textbf{\underline{69.64}} & 
            \underline{84.32} & 
            \textbf{\underline{89.12}} & 
            \textbf{\underline{76.24}} \\
            \bottomrule
            
            
	\end{tabular}
\end{table}

\subsection{Impact of constrained decoding and scaling with training data}
We measure the benefit of applying grounded constrained decoding to using our model without it. We also examine how well the method scales with the availability of training data with and without constrained decoding. We measure the metrics directly on the output Cypher queries, executing the ones with highest probabilities first, until there are 20 nodes. To accurately evaluate LLM $L_1$ and constrained decoding we do not apply LLM $L_2$.

As can be seen in Table~\ref{tab:ablation-constrained}, when constrained decoding is not used and the model is trained on 100\% of available data, it gives slightly lower metrics. When we only use 10\% of the training data, our method only shows a slight decrease in all metrics. However, when used without constrained decoding, the drop becomes larger (e.g 16\% for Recall@20). This suggests that our method is both extremely sample efficient and scales well with more training Q\&As. The advantage of constrained decoding is the most significant when training data is scarce, which is common in any real-world setting. The final row uses the gemma2-9b-text2cypher model without any finetuning on STaRK-prime.
It is not able to out-of-the-box answer any question and applying constrained decoding without finetuning the LLM at all already gives us results close to several baselines.

\begin{table}
	\centering
	\caption{Metrics on STaRK-prime using 10\% of validation data. Percentage of train data used is specified next to the method. Numbers in brackets representing using the method without applying constrained decoding. No schema is provided in prompt and response executed as is in all queries.}
	\label{tab:ablation-constrained}
	\begin{tabular}{c|cccc}
		\toprule
		  Method (\% Training data used) & Hit@1 & Hit@5 & Recall@20 & MRR \\
		\midrule
            Finetuned, 100\%& 44.20(43.75) & 69.20(63.39)& 75.87(69.83) & 0.5528(0.5255) \\
            Finetuned, 10\% & 41.07(35.71) & 66.07(54.91) & 76.11(60.53) & 0.5197(0.4400)\\
            LLM, 0\% & 14.73(0.0) & 23.21(0.0) & 27.53(0.0) & 0.1865(0.0) \\
            \bottomrule
	\end{tabular}
\end{table}

\subsection{The use of query engine}
The use of query engines to optimise query plans on DBs has always been one of the main advantages of DB systems. Table~\ref{tab:query-planner} shows an example executed query plan that is optimal according to the query planner. It first fetches nodes of the type Drug from the node label indexes and then traverses along the ENZYME relationship type. It then filter the joined records with predicates on x1. Afterwards it performs the similar traversal from x2 to x3 along INDICATION and filter on x3. Intuitively, the optimal plan finds the all x2:Drug nodes and filter down twice by joining the two ends.

An alternative valid but suboptimal query plan is provided in Appendix~\ref{ap:query-plan} Table~\ref{tab:bad-query-planner}. It starts with scanning all x3:Disease nodes and traverses towards x2 and then x1. It happens to be suboptimal due to the exploding neighbourhood of Disease nodes along the INDICATION relationship type. The query engine with it's cardinality estimator therefore rules out executing this sequence of operators.

This analysis serves as an example of confirming the advantage of using a query engine. As we have pointed out, any of the existing GraphRAG work that iteratively traverses the graph step-by-step is not able to leverage the query engine since the order of execution is fixed. Therefore, extremely inefficient retrieval (such as the ordering shown in Table~\ref{tab:bad-query-planner}) is possible and unavoidable. 

\begin{table*}
    \centering
    \caption{The optimal execution plan for an example retrieval query: \lstinline|MATCH (x1:GeneOrProtein {name: "CYP3A4"})-[r: ENZYME]-(x2: Drug)-[r2: INDICATION]-(x3: Disease {name: "strongyloidiasis"}) RETURN x2.name|. \texttt{Operator} are executed from the bottom up. \texttt{Details} represent the exact execution parameters. \texttt{Estimated Rows} represent expected rows produced. \texttt{Rows} represent actual rows produced. \texttt{DB Hits} measure the amount of work by the storage engine. Total database access is 103989, total allocated memory is 328 bytes.}
    \label{tab:query-planner}
    \ttfamily 
    \resizebox{\textwidth}{!}{ 
    \begin{tabular}{|l|l|l|l|l|l|}
        \toprule
        \multirow{2}{*}{Operator}         & \multirow{2}{*}{Id}  & \multirow{2}{*}{Details}                   & \multirow{2}{*}{\makecell{Estimated \\ Rows}} & \multirow{2}{*}{Rows}  & \multirow{2}{*}{DB Hits} \\
        & & & & & \\
        \midrule
        +ProduceResults  & 0   & n`x2.name`    & 63              & 4     & 0  \\
        \cmidrule(lr){2-6}
        +Projection      & 1   & x2.name AS `x2.name`    & 63             & 4   & 4     \\
        \cmidrule(lr){2-6}
        |+Filter          & 2   & x3.name = \$autostring\_1 AND x3:Disease   & 63    & 4   & 34872\\
        \cmidrule(lr){2-6}
        |+Expand(All)     & 3   & (x2)-[r2:INDICATION]-(x3)      & 1255    & 17432   & 17432\\
        \cmidrule(lr){2-6}
        |+Filter          & 4   & x1.name = \$autostring\_0 AND x1:GeneOrProtein  & 532                 & 1932   & 25132 \\
        \cmidrule(lr){2-6}
        |+Expand(All)     & 5   & (x2)-[r:ENZYME]-(x1)       & 10634  & 10634   & 18591 \\
        \cmidrule(lr){2-6}
        +NodeByLabelScan  & 6   & x2:Drug               & 7957             & 7957   & 7958 \\
        \bottomrule
    \end{tabular}
    }
\end{table*}

\subsection{Ablation study on choices of LLMs}
We first verify that using an LLM $L_0$ for entity resolution is needed better than simpler methods such as k-nearest-neighbour (kNN) using text embeddings. We measure that by looking at the quality of the best Cypher created from entities identified by an LLM, 2NN and 5NN. Plots showing the quality of entity resolution by various criteria are shown in Appendix~\ref{ap:ablation} Figure~\ref{fig:entity-resolution-ablation}. 

We also study the different choices of base LLMs for $L_1$. In addition to using a general gemma-based text2cypher model, we also finetune a google/gemma2-9b-it and a meta-llama/Llama-3.1-8B-Instruct model. Metrics on STaRK-prime suggest using a already-fintuned general text2cypher model offers some advantage, and gemma2 performs slightly better tham Llama-3.1. Finetuning any of these base models for $L_1$ already achieves beyond SOTA results even without $L_2$. The table is shown in Appendix~\ref{ap:ablation} Table~\ref{tab:ablation-l1}.


\section{Conclusion} \label{sec:conclusion}
In this work, we introduce GraphRAFT, a simple and modular method that leverages graph DBs by retrieving from it using learnt provably correct and optimal Cypher queries. Our experiments show that GraphRAFT consistently achieves beyond SOTA results using smaller LLMs that fit into a single GPU. Our framework can be applied off-the-shelf to any KG in any domain stored in graph DBs. The finetuning process is sample-efficient and scales with more training data.

GraphRAFT is illustrated on graph DBs supporting Cypher. Exactly the same approach can be used for any other graph query language. Our finetuning process requires existing Q\&A set. Future work that addresses these limitations will improve the general applicability of the method.

\bibliographystyle{unsrtnat}
\bibliography{references}

@software{unsloth,
  author = {Daniel Han, Michael Han and Unsloth team},
  title = {Unsloth},
  url = {http://github.com/unslothai/unsloth},
  year = {2023}
}

@article{huang2025survey,
  title={A survey on hallucination in large language models: Principles, taxonomy, challenges, and open questions},
  author={Huang, Lei and Yu, Weijiang and Ma, Weitao and Zhong, Weihong and Feng, Zhangyin and Wang, Haotian and Chen, Qianglong and Peng, Weihua and Feng, Xiaocheng and Qin, Bing and others},
  journal={ACM Transactions on Information Systems},
  volume={43},
  number={2},
  pages={1--55},
  year={2025},
  publisher={ACM New York, NY}
}

@article{lewis2020retrieval,
  title={Retrieval-augmented generation for knowledge-intensive nlp tasks},
  author={Lewis, Patrick and Perez, Ethan and Piktus, Aleksandra and Petroni, Fabio and Karpukhin, Vladimir and Goyal, Naman and K{\"u}ttler, Heinrich and Lewis, Mike and Yih, Wen-tau and Rockt{\"a}schel, Tim and others},
  journal={Advances in neural information processing systems},
  volume={33},
  pages={9459--9474},
  year={2020}
}

@misc{beurerkellner2024guidingllmsrightway,
      title={Guiding LLMs The Right Way: Fast, Non-Invasive Constrained Generation}, 
      author={Luca Beurer-Kellner and Marc Fischer and Martin Vechev},
      year={2024},
      eprint={2403.06988},
      archivePrefix={arXiv},
      primaryClass={cs.LG},
      url={https://arxiv.org/abs/2403.06988}, 
}

@misc{geng2024grammarconstraineddecodingstructurednlp,
      title={Grammar-Constrained Decoding for Structured NLP Tasks without Finetuning}, 
      author={Saibo Geng and Martin Josifoski and Maxime Peyrard and Robert West},
      year={2024},
      eprint={2305.13971},
      archivePrefix={arXiv},
      primaryClass={cs.CL},
      url={https://arxiv.org/abs/2305.13971}, 
}

@article{hu2022lora,
  title={Lora: Low-rank adaptation of large language models.},
  author={Hu, Edward J and Shen, Yelong and Wallis, Phillip and Allen-Zhu, Zeyuan and Li, Yuanzhi and Wang, Shean and Wang, Lu and Chen, Weizhu and others},
  journal={ICLR},
  volume={1},
  number={2},
  pages={3},
  year={2022}
}

@misc{openai,
  key = {OpenAI embeddings},
  howpublished = {\url{https://platform.openai.com/docs/guides/embeddings}},
  year = {2025}
}

@article{grattafiori2024llama,
  title={The llama 3 herd of models},
  author={Grattafiori, Aaron and Dubey, Abhimanyu and Jauhri, Abhinav and Pandey, Abhinav and Kadian, Abhishek and Al-Dahle, Ahmad and Letman, Aiesha and Mathur, Akhil and Schelten, Alan and Vaughan, Alex and others},
  journal={arXiv preprint arXiv:2407.21783},
  year={2024}
}

@article{team2024gemma,
  title={Gemma 2: Improving open language models at a practical size},
  author={Team, Gemma and Riviere, Morgane and Pathak, Shreya and Sessa, Pier Giuseppe and Hardin, Cassidy and Bhupatiraju, Surya and Hussenot, L{\'e}onard and Mesnard, Thomas and Shahriari, Bobak and Ram{\'e}, Alexandre and others},
  journal={arXiv preprint arXiv:2408.00118},
  year={2024}
}

@article{ozsoy2024text2cypher,
  title={Text2Cypher: Bridging Natural Language and Graph Databases},
  author={Ozsoy, Makbule Gulcin and Messallem, Leila and Besga, Jon and Minneci, Gianandrea},
  journal={arXiv preprint arXiv:2412.10064},
  year={2024}
}

@article{edge2024local,
  title={From local to global: A graph rag approach to query-focused summarization},
  author={Edge, Darren and Trinh, Ha and Cheng, Newman and Bradley, Joshua and Chao, Alex and Mody, Apurva and Truitt, Steven and Larson, Jonathan},
  journal={arXiv preprint arXiv:2404.16130},
  year={2024}
}

@inproceedings{raft2024,
  title={RAFT: Adapting Language Model to Domain Specific RAG},
  author={Tianjun Zhang and Shishir G. Patil and Naman Jain and Sheng Shen and Matei Zaharia and Ion Stoica and Joseph E. Gonzalez},
  year={2024},
  journal={arXiv preprint arXiv:2403.10131}
}

@article{li2024simple,
  title={Simple is Effective: The Roles of Graphs and Large Language Models in Knowledge-Graph-Based Retrieval-Augmented Generation},
  author={Li, Mufei and Miao, Siqi and Li, Pan},
  journal={arXiv preprint arXiv:2410.20724},
  year={2024}
}

@inproceedings{luo2024reasoning,
    title={Reasoning on Graphs: Faithful and Interpretable Large Language Model Reasoning},
    author={Linhao Luo and Yuan-Fang Li and Reza Haf and Shirui Pan},
    booktitle={The Twelfth International Conference on Learning Representations},
    year={2024},
    url={https://openreview.net/forum?id=ZGNWW7xZ6Q}
}

@inproceedings{sun2024thinkongraph,
title={Think-on-Graph: Deep and Responsible Reasoning of Large Language Model on Knowledge Graph},
author={Jiashuo Sun and Chengjin Xu and Lumingyuan Tang and Saizhuo Wang and Chen Lin and Yeyun Gong and Lionel Ni and Heung-Yeung Shum and Jian Guo},
booktitle={The Twelfth International Conference on Learning Representations},
year={2024},
url={https://openreview.net/forum?id=nnVO1PvbTv}
}

@article{ma2024think,
  title={Think-on-Graph 2.0: Deep and Faithful Large Language Model Reasoning with Knowledge-guided Retrieval Augmented Generation},
  author={Ma, Shengjie and Xu, Chengjin and Jiang, Xuhui and Li, Muzhi and Qu, Huaren and Yang, Cehao and Mao, Jiaxin and Guo, Jian},
  journal={arXiv preprint arXiv:2407.10805},
  year={2024}
}

@misc{hu2024graggraphretrievalaugmentedgeneration,
      title={GRAG: Graph Retrieval-Augmented Generation}, 
      author={Yuntong Hu and Zhihan Lei and Zheng Zhang and Bo Pan and Chen Ling and Liang Zhao},
      year={2024},
      eprint={2405.16506},
      archivePrefix={arXiv},
      primaryClass={cs.LG},
      url={https://arxiv.org/abs/2405.16506}, 
}

@misc{sui2024fidelisfaithfulreasoninglarge,
      title={FiDeLiS: Faithful Reasoning in Large Language Model for Knowledge Graph Question Answering}, 
      author={Yuan Sui and Yufei He and Nian Liu and Xiaoxin He and Kun Wang and Bryan Hooi},
      year={2024},
      eprint={2405.13873},
      archivePrefix={arXiv},
      primaryClass={cs.AI},
      url={https://arxiv.org/abs/2405.13873}, 
}

@inproceedings{
wang2025reasoning,
title={Reasoning of Large Language Models over Knowledge Graphs with Super-Relations},
author={Song Wang and Junhong Lin and Xiaojie Guo and Julian Shun and Jundong Li and Yada Zhu},
booktitle={The Thirteenth International Conference on Learning Representations},
year={2025},
url={https://openreview.net/forum?id=rTCJ29pkuA}
}

@misc{wang2023knowledgedrivencotexploringfaithful,
      title={Knowledge-Driven CoT: Exploring Faithful Reasoning in LLMs for Knowledge-intensive Question Answering}, 
      author={Keheng Wang and Feiyu Duan and Sirui Wang and Peiguang Li and Yunsen Xian and Chuantao Yin and Wenge Rong and Zhang Xiong},
      year={2023},
      eprint={2308.13259},
      archivePrefix={arXiv},
      primaryClass={cs.CL},
      url={https://arxiv.org/abs/2308.13259}, 
}

@inproceedings{Xu_2024, 
   series={SIGIR 2024},
   title={Retrieval-Augmented Generation with Knowledge Graphs for Customer Service Question Answering},
   url={http://dx.doi.org/10.1145/3626772.3661370},
   DOI={10.1145/3626772.3661370},
   booktitle={Proceedings of the 47th International ACM SIGIR Conference on Research and Development in Information Retrieval},
   publisher={ACM},
   author={Xu, Zhentao and Cruz, Mark Jerome and Guevara, Matthew and Wang, Tie and Deshpande, Manasi and Wang, Xiaofeng and Li, Zheng},
   year={2024},
   month=jul, pages={2905–2909},
   collection={SIGIR 2024} }

@misc{graft,
      title={GRAFT: Graph Retrieval Augmented Fine Tuning for Multi-Hop Query Summarization}, 
      author={Sonya Jin and Sunny Yu and Natalia Kokoromyti},
      year={2025},
      url={https://web.stanford.edu/class/cs224n/final-reports/256724569.pdf}, 
}

@article{lan2021survey,
  title={A survey on complex knowledge base question answering: Methods, challenges and solutions},
  author={Lan, Yunshi and He, Gaole and Jiang, Jinhao and Jiang, Jing and Zhao, Wayne Xin and Wen, Ji-Rong},
  journal={arXiv preprint arXiv:2105.11644},
  year={2021}
}

@misc{baek2023knowledgeaugmentedlanguagemodelprompting,
      title={Knowledge-Augmented Language Model Prompting for Zero-Shot Knowledge Graph Question Answering}, 
      author={Jinheon Baek and Alham Fikri Aji and Amir Saffari},
      year={2023},
      eprint={2306.04136},
      archivePrefix={arXiv},
      primaryClass={cs.CL},
      url={https://arxiv.org/abs/2306.04136}, 
}

@article{Robertson2009ThePR,
  title={The Probabilistic Relevance Framework: BM25 and Beyond},
  author={Stephen E. Robertson and Hugo Zaragoza},
  journal={Found. Trends Inf. Retr.},
  year={2009},
  volume={3},
  pages={333-389},
  url={https://api.semanticscholar.org/CorpusID:207178704}
}

@Misc{voyageai,
author =   {Voyage AI},
title =    {Voyage AI Text embedding models},
howpublished = {\url{https://docs.voyageai.com/reference/embeddings-api}},
year = {2009}
}

@inproceedings{muennighoff2024generative,
  title={Generative representational instruction tuning},
  author={Muennighoff, Niklas and Hongjin, SU and Wang, Liang and Yang, Nan and Wei, Furu and Yu, Tao and Singh, Amanpreet and Kiela, Douwe},
  booktitle={ICLR 2024 Workshop: How Far Are We From AGI},
  year={2024}
}

@inproceedings{santhanam-etal-2022-colbertv2,
    title = "{C}ol{BERT}v2: Effective and Efficient Retrieval via Lightweight Late Interaction",
    author = "Santhanam, Keshav  and
      Khattab, Omar  and
      Saad-Falcon, Jon  and
      Potts, Christopher  and
      Zaharia, Matei",
    editor = "Carpuat, Marine  and
      de Marneffe, Marie-Catherine  and
      Meza Ruiz, Ivan Vladimir",
    booktitle = "Proceedings of the 2022 Conference of the North American Chapter of the Association for Computational Linguistics: Human Language Technologies",
    month = jul,
    year = "2022",
    address = "Seattle, United States",
    publisher = "Association for Computational Linguistics",
    url = "https://aclanthology.org/2022.naacl-main.272/",
    doi = "10.18653/v1/2022.naacl-main.272",
    pages = "3715--3734",
}

@article{wu2025avatar,
  title={AvaTaR: Optimizing LLM Agents for Tool Usage via Contrastive Reasoning},
  author={Wu, Shirley and Zhao, Shiyu and Huang, Qian and Huang, Kexin and Yasunaga, Michihiro and Cao, Kaidi and Ioannidis, Vassilis and Subbian, Karthik and Leskovec, Jure and Zou, James Y},
  journal={Advances in Neural Information Processing Systems},
  volume={37},
  pages={25981--26010},
  year={2025}
}

@misc{lee2024hybgraghybridretrievalaugmentedgeneration,
      title={HybGRAG: Hybrid Retrieval-Augmented Generation on Textual and Relational Knowledge Bases}, 
      author={Meng-Chieh Lee and Qi Zhu and Costas Mavromatis and Zhen Han and Soji Adeshina and Vassilis N. Ioannidis and Huzefa Rangwala and Christos Faloutsos},
      year={2024},
      eprint={2412.16311},
      archivePrefix={arXiv},
      primaryClass={cs.LG},
      url={https://arxiv.org/abs/2412.16311}, 
}

@misc{lei2025mixturestructuralandtextualretrievaltextrich,
      title={Mixture of Structural-and-Textual Retrieval over Text-rich Graph Knowledge Bases}, 
      author={Yongjia Lei and Haoyu Han and Ryan A. Rossi and Franck Dernoncourt and Nedim Lipka and Mahantesh M Halappanavar and Jiliang Tang and Yu Wang},
      year={2025},
      eprint={2502.20317},
      archivePrefix={arXiv},
      primaryClass={cs.LG},
      url={https://arxiv.org/abs/2502.20317}, 
}

@misc{xia2025knowledgeawarequeryexpansionlarge,
      title={Knowledge-Aware Query Expansion with Large Language Models for Textual and Relational Retrieval}, 
      author={Yu Xia and Junda Wu and Sungchul Kim and Tong Yu and Ryan A. Rossi and Haoliang Wang and Julian McAuley},
      year={2025},
      eprint={2410.13765},
      archivePrefix={arXiv},
      primaryClass={cs.CL},
      url={https://arxiv.org/abs/2410.13765}, 
}

@misc{li2024multifieldadaptiveretrieval,
      title={Multi-Field Adaptive Retrieval}, 
      author={Millicent Li and Tongfei Chen and Benjamin Van Durme and Patrick Xia},
      year={2024},
      eprint={2410.20056},
      archivePrefix={arXiv},
      primaryClass={cs.IR},
      url={https://arxiv.org/abs/2410.20056}, 
}

@inproceedings{
chen2025knowledge,
title={Knowledge Graph Finetuning Enhances Knowledge Manipulation in Large Language Models},
author={Hanzhu Chen and Xu Shen and Jie Wang and Zehao Wang and Qitan Lv and Junjie He and Rong Wu and Feng Wu and Jieping Ye},
booktitle={The Thirteenth International Conference on Learning Representations},
year={2025},
url={https://openreview.net/forum?id=oMFOKjwaRS}
}

@inproceedings{
he2024gretriever,
title={G-Retriever: Retrieval-Augmented Generation for Textual Graph Understanding and Question Answering},
author={Xiaoxin He and Yijun Tian and Yifei Sun and Nitesh V Chawla and Thomas Laurent and Yann LeCun and Xavier Bresson and Bryan Hooi},
booktitle={The Thirty-eighth Annual Conference on Neural Information Processing Systems},
year={2024},
url={https://openreview.net/forum?id=MPJ3oXtTZl}
}

@article{mavromatis2024gnn,
  title={GNN-RAG: Graph Neural Retrieval for Large Language Model Reasoning},
  author={Mavromatis, Costas and Karypis, George},
  journal={arXiv preprint arXiv:2405.20139},
  year={2024}
}

@inproceedings{zhao2023learning,
    title={Learning on Large-scale Text-attributed Graphs via Variational Inference},
    author={Jianan Zhao and Meng Qu and Chaozhuo Li and Hao Yan and Qian Liu and Rui Li and Xing Xie and Jian Tang},
    booktitle={The Eleventh International Conference on Learning Representations },
    year={2023},
    url={https://openreview.net/forum?id=q0nmYciuuZN}
}

@article{zhang2022greaselm,
  title={Greaselm: Graph reasoning enhanced language models for question answering},
  author={Zhang, Xikun and Bosselut, Antoine and Yasunaga, Michihiro and Ren, Hongyu and Liang, Percy and Manning, Christopher D and Leskovec, Jure},
  journal={arXiv preprint arXiv:2201.08860},
  year={2022}
}

@article{yasunaga2021qa,
  title={QA-GNN: Reasoning with language models and knowledge graphs for question answering},
  author={Yasunaga, Michihiro and Ren, Hongyu and Bosselut, Antoine and Liang, Percy and Leskovec, Jure},
  journal={arXiv preprint arXiv:2104.06378},
  year={2021}
}

@inproceedings{
jiang2024ragraph,
title={{RAG}raph: A General Retrieval-Augmented Graph Learning Framework},
author={Xinke Jiang and Rihong Qiu and Yongxin Xu and WentaoZhang and Yichen Zhu and Ruizhe zhang and Yuchen Fang and Xu Chu and Junfeng Zhao and Yasha Wang},
booktitle={The Thirty-eighth Annual Conference on Neural Information Processing Systems},
year={2024},
url={https://openreview.net/forum?id=Dzk2cRUFMt}
}

@inproceedings{
he2024harnessing,
title={Harnessing Explanations: {LLM}-to-{LM} Interpreter for Enhanced Text-Attributed Graph Representation Learning},
author={Xiaoxin He and Xavier Bresson and Thomas Laurent and Adam Perold and Yann LeCun and Bryan Hooi},
booktitle={The Twelfth International Conference on Learning Representations},
year={2024},
url={https://openreview.net/forum?id=RXFVcynVe1}
}

@misc{tokengt,
      title={Pure Transformers are Powerful Graph Learners}, 
      author={Jinwoo Kim and Tien Dat Nguyen and Seonwoo Min and Sungjun Cho and Moontae Lee and Honglak Lee and Seunghoon Hong},
      year={2022},
      eprint={2207.02505},
      archivePrefix={arXiv},
      primaryClass={cs.LG},
      url={https://arxiv.org/abs/2207.02505}, 
}

@misc{ying2021transformersreallyperformbad,
      title={Do Transformers Really Perform Bad for Graph Representation?}, 
      author={Chengxuan Ying and Tianle Cai and Shengjie Luo and Shuxin Zheng and Guolin Ke and Di He and Yanming Shen and Tie-Yan Liu},
      year={2021},
      eprint={2106.05234},
      archivePrefix={arXiv},
      primaryClass={cs.LG},
      url={https://arxiv.org/abs/2106.05234}, 
}

@misc{lei2025spider20evaluatinglanguage,
      title={Spider 2.0: Evaluating Language Models on Real-World Enterprise Text-to-SQL Workflows}, 
      author={Fangyu Lei and Jixuan Chen and Yuxiao Ye and Ruisheng Cao and Dongchan Shin and Hongjin Su and Zhaoqing Suo and Hongcheng Gao and Wenjing Hu and Pengcheng Yin and Victor Zhong and Caiming Xiong and Ruoxi Sun and Qian Liu and Sida Wang and Tao Yu},
      year={2025},
      eprint={2411.07763},
      archivePrefix={arXiv},
      primaryClass={cs.CL},
      url={https://arxiv.org/abs/2411.07763}, 
}

@INPROCEEDINGS{texttosparql,
  author={Avila, Caio Viktor S. and Vidal, Vânia M.P. and Franco, Wellington and Casanova, Marco A.},
  booktitle={2024 IEEE 18th International Conference on Semantic Computing (ICSC)}, 
  title={Experiments with text-to-SPARQL based on ChatGPT}, 
  year={2024},
  volume={},
  number={},
  pages={277-284},
  doi={10.1109/ICSC59802.2024.00050}}

@misc{brei2024leveragingsmalllanguagemodels,
      title={Leveraging small language models for Text2SPARQL tasks to improve the resilience of AI assistance}, 
      author={Felix Brei and Johannes Frey and Lars-Peter Meyer},
      year={2024},
      eprint={2405.17076},
      archivePrefix={arXiv},
      primaryClass={cs.AI},
      url={https://arxiv.org/abs/2405.17076}, 
}

@misc{ozsoy2024text2cypherbridgingnaturallanguage,
      title={Text2Cypher: Bridging Natural Language and Graph Databases}, 
      author={Makbule Gulcin Ozsoy and Leila Messallem and Jon Besga and Gianandrea Minneci},
      year={2024},
      eprint={2412.10064},
      archivePrefix={arXiv},
      primaryClass={cs.LG},
      url={https://arxiv.org/abs/2412.10064}, 
}

@inproceedings{yih-etal-2015-semantic,
    title = "Semantic Parsing via Staged Query Graph Generation: Question Answering with Knowledge Base",
    author = "Yih, Wen-tau  and
      Chang, Ming-Wei  and
      He, Xiaodong  and
      Gao, Jianfeng",
    editor = "Zong, Chengqing  and
      Strube, Michael",
    booktitle = "Proceedings of the 53rd Annual Meeting of the Association for Computational Linguistics and the 7th International Joint Conference on Natural Language Processing (Volume 1: Long Papers)",
    month = jul,
    year = "2015",
    address = "Beijing, China",
    publisher = "Association for Computational Linguistics",
    url = "https://aclanthology.org/P15-1128",
    doi = "10.3115/v1/P15-1128",
    pages = "1321--1331",
}

@misc{xia2024knowledgeawarequeryexpansionlarge,
      title={Knowledge-Aware Query Expansion with Large Language Models for Textual and Relational Retrieval}, 
      author={Yu Xia and Junda Wu and Sungchul Kim and Tong Yu and Ryan A. Rossi and Haoliang Wang and Julian McAuley},
      year={2024},
      eprint={2410.13765},
      archivePrefix={arXiv},
      primaryClass={cs.CL},
      url={https://arxiv.org/abs/2410.13765}, 
}

@inproceedings{luo-etal-2018-knowledge,
    title = "Knowledge Base Question Answering via Encoding of Complex Query Graphs",
    author = "Luo, Kangqi  and
      Lin, Fengli  and
      Luo, Xusheng  and
      Zhu, Kenny",
    editor = "Riloff, Ellen  and
      Chiang, David  and
      Hockenmaier, Julia  and
      Tsujii, Jun{'}ichi",
    booktitle = "Proceedings of the 2018 Conference on Empirical Methods in Natural Language Processing",
    month = oct # "-" # nov,
    year = "2018",
    address = "Brussels, Belgium",
    publisher = "Association for Computational Linguistics",
    url = "https://aclanthology.org/D18-1242/",
    doi = "10.18653/v1/D18-1242",
    pages = "2185--2194"
}

@inproceedings{10.1007/978-3-319-93417-4_46,
author = {Zafar, Hamid and Napolitano, Giulio and Lehmann, Jens},
title = {Formal Query Generation for Question Answering over Knowledge Bases},
year = {2018},
isbn = {978-3-319-93416-7},
publisher = {Springer-Verlag},
address = {Berlin, Heidelberg},
url = {https://doi.org/10.1007/978-3-319-93417-4_46},
doi = {10.1007/978-3-319-93417-4_46},
booktitle = {The Semantic Web: 15th International Conference, ESWC 2018, Heraklion, Crete, Greece, June 3–7, 2018, Proceedings},
pages = {714–728},
numpages = {15},
location = {Heraklion, Greece}
}

@inproceedings{10.1145/2806416.2806472,
author = {Bast, Hannah and Haussmann, Elmar},
title = {More Accurate Question Answering on Freebase},
year = {2015},
isbn = {9781450337946},
publisher = {Association for Computing Machinery},
address = {New York, NY, USA},
url = {https://doi.org/10.1145/2806416.2806472},
doi = {10.1145/2806416.2806472},
booktitle = {Proceedings of the 24th ACM International on Conference on Information and Knowledge Management},
pages = {1431–1440},
numpages = {10},
location = {Melbourne, Australia},
series = {CIKM '15}
}

@inproceedings{chen-etal-2019-uhop,
    title = "{UH}op: An Unrestricted-Hop Relation Extraction Framework for Knowledge-Based Question Answering",
    author = "Chen, Zi-Yuan  and
      Chang, Chih-Hung  and
      Chen, Yi-Pei  and
      Nayak, Jijnasa  and
      Ku, Lun-Wei",
    editor = "Burstein, Jill  and
      Doran, Christy  and
      Solorio, Thamar",
    booktitle = "Proceedings of the 2019 Conference of the North {A}merican Chapter of the Association for Computational Linguistics: Human Language Technologies, Volume 1 (Long and Short Papers)",
    month = jun,
    year = "2019",
    address = "Minneapolis, Minnesota",
    publisher = "Association for Computational Linguistics",
    url = "https://aclanthology.org/N19-1031/",
    doi = "10.18653/v1/N19-1031",
    pages = "345--356"
}

@inproceedings{10.1145/3357384.3358033,
author = {Bhutani, Nikita and Zheng, Xinyi and Jagadish, H V},
title = {Learning to Answer Complex Questions over Knowledge Bases with Query Composition},
year = {2019},
isbn = {9781450369763},
publisher = {Association for Computing Machinery},
address = {New York, NY, USA},
url = {https://doi.org/10.1145/3357384.3358033},
doi = {10.1145/3357384.3358033},
booktitle = {Proceedings of the 28th ACM International Conference on Information and Knowledge Management},
pages = {739–748},
numpages = {10},
location = {Beijing, China},
series = {CIKM '19}
}

@inproceedings{lan-jiang-2020-query,
    title = "Query Graph Generation for Answering Multi-hop Complex Questions from Knowledge Bases",
    author = "Lan, Yunshi  and
      Jiang, Jing",
    editor = "Jurafsky, Dan  and
      Chai, Joyce  and
      Schluter, Natalie  and
      Tetreault, Joel",
    booktitle = "Proceedings of the 58th Annual Meeting of the Association for Computational Linguistics",
    month = jul,
    year = "2020",
    address = "Online",
    publisher = "Association for Computational Linguistics",
    url = "https://aclanthology.org/2020.acl-main.91/",
    doi = "10.18653/v1/2020.acl-main.91",
    pages = "969--974"
}

@misc{sun2019pullnetopendomainquestion,
      title={PullNet: Open Domain Question Answering with Iterative Retrieval on Knowledge Bases and Text}, 
      author={Haitian Sun and Tania Bedrax-Weiss and William W. Cohen},
      year={2019},
      eprint={1904.09537},
      archivePrefix={arXiv},
      primaryClass={cs.CL},
      url={https://arxiv.org/abs/1904.09537}, 
}

@inproceedings{sun-etal-2018-open,
    title = "Open Domain Question Answering Using Early Fusion of Knowledge Bases and Text",
    author = "Sun, Haitian  and
      Dhingra, Bhuwan  and
      Zaheer, Manzil  and
      Mazaitis, Kathryn  and
      Salakhutdinov, Ruslan  and
      Cohen, William",
    editor = "Riloff, Ellen  and
      Chiang, David  and
      Hockenmaier, Julia  and
      Tsujii, Jun{'}ichi",
    booktitle = "Proceedings of the 2018 Conference on Empirical Methods in Natural Language Processing",
    month = oct # "-" # nov,
    year = "2018",
    address = "Brussels, Belgium",
    publisher = "Association for Computational Linguistics",
    url = "https://aclanthology.org/D18-1455/",
    doi = "10.18653/v1/D18-1455",
    pages = "4231--4242"
}

@inproceedings{xiong-etal-2019-improving,
    title = "Improving Question Answering over Incomplete {KB}s with Knowledge-Aware Reader",
    author = "Xiong, Wenhan  and
      Yu, Mo  and
      Chang, Shiyu  and
      Guo, Xiaoxiao  and
      Wang, William Yang",
    editor = "Korhonen, Anna  and
      Traum, David  and
      M{\`a}rquez, Llu{\'i}s",
    booktitle = "Proceedings of the 57th Annual Meeting of the Association for Computational Linguistics",
    month = jul,
    year = "2019",
    address = "Florence, Italy",
    publisher = "Association for Computational Linguistics",
    url = "https://aclanthology.org/P19-1417/",
    doi = "10.18653/v1/P19-1417",
    pages = "4258--4264"
}

@inproceedings{he-etal-2017-generating,
    title = "Generating Natural Answers by Incorporating Copying and Retrieving Mechanisms in Sequence-to-Sequence Learning",
    author = "He, Shizhu  and
      Liu, Cao  and
      Liu, Kang  and
      Zhao, Jun",
    editor = "Barzilay, Regina  and
      Kan, Min-Yen",
    booktitle = "Proceedings of the 55th Annual Meeting of the Association for Computational Linguistics (Volume 1: Long Papers)",
    month = jul,
    year = "2017",
    address = "Vancouver, Canada",
    publisher = "Association for Computational Linguistics",
    url = "https://aclanthology.org/P17-1019/",
    doi = "10.18653/v1/P17-1019",
    pages = "199--208"
}

@inproceedings{freebase,
author = {Bollacker, Kurt and Evans, Colin and Paritosh, Praveen and Sturge, Tim and Taylor, Jamie},
title = {Freebase: a collaboratively created graph database for structuring human knowledge},
year = {2008},
isbn = {9781605581026},
publisher = {Association for Computing Machinery},
address = {New York, NY, USA},
url = {https://doi.org/10.1145/1376616.1376746},
doi = {10.1145/1376616.1376746},
booktitle = {Proceedings of the 2008 ACM SIGMOD International Conference on Management of Data},
pages = {1247–1250},
numpages = {4},
location = {Vancouver, Canada},
series = {SIGMOD '08}
}

@inproceedings{Yih2016TheVO,
  title={The Value of Semantic Parse Labeling for Knowledge Base Question Answering},
  author={Wen-tau Yih and Matthew Richardson and Christopher Meek and Ming-Wei Chang and Jina Suh},
  booktitle={Annual Meeting of the Association for Computational Linguistics},
  year={2016},
  url={https://api.semanticscholar.org/CorpusID:13905064}
}

@inproceedings{talmor-berant-2018-web,
    title = "The Web as a Knowledge-Base for Answering Complex Questions",
    author = "Talmor, Alon  and
      Berant, Jonathan",
    editor = "Walker, Marilyn  and
      Ji, Heng  and
      Stent, Amanda",
    booktitle = "Proceedings of the 2018 Conference of the North {A}merican Chapter of the Association for Computational Linguistics: Human Language Technologies, Volume 1 (Long Papers)",
    month = jun,
    year = "2018",
    address = "New Orleans, Louisiana",
    publisher = "Association for Computational Linguistics",
    url = "https://aclanthology.org/N18-1059",
    doi = "10.18653/v1/N18-1059",
    pages = "641--651",
}

@inproceedings{qi2021answering,
  title={Answering Open-Domain Questions of Varying Reasoning Steps from Text},
  author = {Qi, Peng and Lee, Haejun and Sido, Oghenetegiri "TG" and Manning, Christopher D.},
  booktitle = {Empirical Methods for Natural Language Processing ({EMNLP})},
  year = {2021}
}

@inproceedings{yang2018hotpotqa,
  title={{HotpotQA}: A Dataset for Diverse, Explainable Multi-hop Question Answering},
  author={Yang, Zhilin and Qi, Peng and Zhang, Saizheng and Bengio, Yoshua and Cohen, William W. and Salakhutdinov, Ruslan and Manning, Christopher D.},
  booktitle={Conference on Empirical Methods in Natural Language Processing ({EMNLP})},
  year={2018}
}

@article{wu2025stark,
  title={Stark: Benchmarking llm retrieval on textual and relational knowledge bases},
  author={Wu, Shirley and Zhao, Shiyu and Yasunaga, Michihiro and Huang, Kexin and Cao, Kaidi and Huang, Qian and Ioannidis, Vassilis and Subbian, Karthik and Zou, James Y and Leskovec, Jure},
  journal={Advances in Neural Information Processing Systems},
  volume={37},
  pages={127129--127153},
  year={2025}
}

@article{chandak2023building,
  title={Building a knowledge graph to enable precision medicine},
  author={Chandak, Payal and Huang, Kexin and Zitnik, Marinka},
  journal={Scientific Data},
  volume={10},
  number={1},
  pages={67},
  url={https://doi.org/10.1038/s41597-023-01960-3},
  year={2023},
  publisher={Nature Publishing Group}
}

@article{hu2020open,
  title={Open graph benchmark: Datasets for machine learning on graphs},
  author={Hu, Weihua and Fey, Matthias and Zitnik, Marinka and Dong, Yuxiao and Ren, Hongyu and Liu, Bowen and Catasta, Michele and Leskovec, Jure},
  journal={Advances in neural information processing systems},
  volume={33},
  pages={22118--22133},
  year={2020}
}

@article{bechler2025position,
  title={Position: Graph Learning Will Lose Relevance Due To Poor Benchmarks},
  author={Bechler-Speicher, Maya and Finkelshtein, Ben and Frasca, Fabrizio and M{\"u}ller, Luis and T{\"o}nshoff, Jan and Siraudin, Antoine and Zaverkin, Viktor and Bronstein, Michael M and Niepert, Mathias and Perozzi, Bryan and others},
  journal={arXiv preprint arXiv:2502.14546},
  year={2025}
}

@inproceedings{Berant2013SemanticPO,
  title={Semantic Parsing on Freebase from Question-Answer Pairs},
  author={Jonathan Berant and Andrew K. Chou and Roy Frostig and Percy Liang},
  booktitle={Conference on Empirical Methods in Natural Language Processing},
  year={2013},
  url={https://api.semanticscholar.org/CorpusID:6401679}
}

@inproceedings{gu2021beyond,
  title={Beyond IID: three levels of generalization for question answering on knowledge bases},
  author={Gu, Yu and Kase, Sue and Vanni, Michelle and Sadler, Brian and Liang, Percy and Yan, Xifeng and Su, Yu},
  booktitle={Proceedings of the Web Conference 2021},
  pages={3477--3488},
  year={2021}
}

@article{wikidata,
author = {Vrande\v{c}i\'{c}, Denny and Kr\"{o}tzsch, Markus},
title = {Wikidata: a free collaborative knowledgebase},
year = {2014},
issue_date = {October 2014},
publisher = {Association for Computing Machinery},
address = {New York, NY, USA},
volume = {57},
number = {10},
issn = {0001-0782},
url = {https://doi.org/10.1145/2629489},
doi = {10.1145/2629489},
journal = {Commun. ACM},
month = sep,
pages = {78–85},
numpages = {8}
}

@misc{rdf,
  key = {RDF},
  howpublished = {\url{https://www.w3.org/RDF/}},
  year = {2014}
}

@misc{sparql,
  key = {SPARQL},
  howpublished = {\url{https://www.w3.org/TR/sparql11-query/}},
  year = {2013}
}

@misc{owl,
  key = {OWL},
  howpublished = {\url{https://www.w3.org/TR/owl2-overview/}},
  year = {2012}
}

@misc{gql,
  key = {GQL},
  howpublished = {\url{https://www.iso.org/standard/76120.html}},
  year = {2024}
}

@article{gqlpaper,
  author       = {Alin Deutsch and
                  Nadime Francis and
                  Alastair Green and
                  Keith Hare and
                  Bei Li and
                  Leonid Libkin and
                  Tobias Lindaaker and
                  Victor Marsault and
                  Wim Martens and
                  Jan Michels and
                  Filip Murlak and
                  Stefan Plantikow and
                  Petra Selmer and
                  Hannes Voigt and
                  Oskar van Rest and
                  Domagoj Vrgoc and
                  Mingxi Wu and
                  Fred Zemke},
  title        = {Graph Pattern Matching in {GQL} and {SQL/PGQ}},
  journal      = {CoRR},
  volume       = {abs/2112.06217},
  year         = {2021},
  url          = {https://arxiv.org/abs/2112.06217},
  eprinttype    = {arXiv},
  eprint       = {2112.06217},
  bibsource    = {dblp computer science bibliography, https://dblp.org}
}

@article{cyphersemantics,
  author       = {Nadime Francis and
                  Alastair Green and
                  Paolo Guagliardo and
                  Leonid Libkin and
                  Tobias Lindaaker and
                  Victor Marsault and
                  Stefan Plantikow and
                  Mats Rydberg and
                  Martin Schuster and
                  Petra Selmer and
                  Andr{\'{e}}s Taylor},
  title        = {Formal Semantics of the Language Cypher},
  journal      = {CoRR},
  volume       = {abs/1802.09984},
  year         = {2018},
  url          = {http://arxiv.org/abs/1802.09984},
  eprinttype    = {arXiv},
  eprint       = {1802.09984},
  bibsource    = {dblp computer science bibliography, https://dblp.org}
}

@inproceedings{cypher,
  author       = {Nadime Francis and
                  Alastair Green and
                  Paolo Guagliardo and
                  Leonid Libkin and
                  Tobias Lindaaker and
                  Victor Marsault and
                  Stefan Plantikow and
                  Mats Rydberg and
                  Petra Selmer and
                  Andr{\'{e}}s Taylor},
  editor       = {Gautam Das and
                  Christopher M. Jermaine and
                  Philip A. Bernstein},
  title        = {Cypher: An Evolving Query Language for Property Graphs},
  booktitle    = {Proceedings of the 2018 International Conference on Management of
                  Data, {SIGMOD} Conference 2018, Houston, TX, USA, June 10-15, 2018},
  pages        = {1433--1445},
  publisher    = {{ACM}},
  year         = {2018},
  url          = {https://doi.org/10.1145/3183713.3190657},
  doi          = {10.1145/3183713.3190657},
  bibsource    = {dblp computer science bibliography, https://dblp.org}
}

@inproceedings{opencypher,
  author       = {Alastair Green and
                  Martin Junghanns and
                  Max Kie{\ss}ling and
                  Tobias Lindaaker and
                  Stefan Plantikow and
                  Petra Selmer},
  editor       = {Michael H. B{\"{o}}hlen and
                  Reinhard Pichler and
                  Norman May and
                  Erhard Rahm and
                  Shan{-}Hung Wu and
                  Katja Hose},
  title        = {openCypher: New Directions in Property Graph Querying},
  booktitle    = {Proceedings of the 21st International Conference on Extending Database
                  Technology, {EDBT} 2018, Vienna, Austria, March 26-29, 2018},
  pages        = {520--523},
  publisher    = {OpenProceedings.org},
  year         = {2018},
  url          = {https://doi.org/10.5441/002/edbt.2018.62},
  doi          = {10.5441/002/EDBT.2018.62},
  bibsource    = {dblp computer science bibliography, https://dblp.org}
}

@misc{neo4j,
  key = {Neo4j},
  howpublished = {\url{https://neo4j.com/}}
}

@misc{arangoDB,
  key = {ArangoDB},
  howpublished = {\url{https://arangodb.com/}}
}

@misc{tigergraph,
  key = {TigerGraph},
  howpublished = {\url{https://www.tigergraph.com/}}
}

@misc{neptune,
  key = {AWS Neptune},
  howpublished = {\url{https://aws.amazon.com/neptune/}}
}


\appendix

\section{Constrained decoding} \label{ap:cd}
We formally state the informal Lemma~\ref{lemma:cd} as three separate formal statements and prove them.

Recall notations in Section~\ref{sec:cd} where $\mathcal{Q} = \{\mathcal{Q}^1, \cdots, \mathcal{Q}^M\}$ is the set of $M$ valid queries for a question, each has tokenization $\mathcal{Q}^k = (n_0, \cdots, n_{Q_k}) \in \mathcal{V}^{Q_k}$ of variable length. We use $l$ and $p$ to denote regular logits and probabilities and $\tilde{l}$ and $\tilde{p}$ to denote those produced by constrained decoding.

{\renewcommand{\thelemma}{1.1}%
\begin{lemma} \label{lemma:cd-1.1}
If $Q \notin \mathcal{Q}$ and $Q$ has tokenization $(t_0, \dots, t_q)$, then $\tilde{p}(Q) = \tilde{p}(t_q | t_{q-1},\dots,t_0) \times \dots  \times \tilde{p}(t_0) = 0$.
\end{lemma}
}

\begin{proof}
If $t_0$ is not a valid first token, $\mathcal{Q}^k_0 \neq t_0$ for all $k \in [1,M]$. Let the logits for the first token produced by LLM $L_1$ be $(l_0, \dots, l_\mathcal{V})$. Let the logits after applying constrained decoding be $(\tilde{l}_0, \dots, \tilde{l}_\mathcal{V})$. Then $\tilde{l}_{n_0} = -\infty$ and the probability of generating token $t_0$ is $\tilde{p}(n_0) = 0$.

Now let $(t_0, \dots, t_i)$ be the longest subsequence of valid tokens for $Q$. This means there are valid queries $Q^{k_0}, \dots, Q^{k_n}$ such that $Q^{k_j}_{0,i} = (t_0, \dots, t_i)$ for $j \in [0,n]$. Since this is the longest subsequence, $Q^{k_j}_{i+1} \neq t_{i+1}$. For any $j$, let the logits of $i+1^{\text{th}}$ token be $(l_0, \dots, l_\mathcal{V})$ and $(\tilde{l}_0, \dots, \tilde{l}_\mathcal{V})$ after constrained decoding. $\tilde{l}_{t_{i+1}} = -\infty$. Hence $\tilde{p}(t_{i+1} | t_i,\dots,t_0) = 0$. Therefore, $\tilde{p}(Q) = \tilde{p}(t_q | t_{q-1},\dots,t_0) \times \dots  \times \tilde{p}(n_0) = 0$ in all cases.
\end{proof}

{\renewcommand{\thelemma}{1.2}%
\begin{lemma} \label{lemma:cd-1.2}
Suppose that LLM $L_1$ has generated a sequence of valid tokens $(t_0, \dots, t_k)$, logits for the next token $k+1$ are $(l_0, \dots, l_\mathcal{V})$. If $t_{k+1}$ and $t'_{k+1}$ are two valid next tokens and $l_{t_{k+1}} > l_{t'_{k+1}}$, then $\tilde{p}(t_{k+1}|t_k,\dots,t_0) > \tilde{p}(t'_{k+1}|t_k,\dots,t_0)$.
\end{lemma}
}

\begin{proof}
Since the two next tokens are valid, $l_{t_{k+1}} = \tilde{l}_{t_{k+1}}$ and $l_{t'_{k+1}} = \tilde{l}_{t'_{k+1}}$. $\tilde{p}(t_{k+1}|t_k,\dots,t_0) = \text{softmax}(\tilde{l}_0, \dots, \tilde{l}_\mathcal{V})_{t_{k+1}}
= e^{\tilde{l}_{t_{k+1}}}/\sum_{i=0}^\mathcal{V} e^{\tilde{l}_i} > e^{\tilde{l}_{t'_{k+1}}}/\sum_{i=0}^\mathcal{V} e^{\tilde{l}_i} = \tilde{p}(t'_{k+1}|t_k,\dots,t_0)$ simply by the monotonicity of the exponential function.
\end{proof}

{\renewcommand{\thelemma}{1.3}%
\begin{lemma} \label{lemma:cd-1.3}
When beam width = M, LLM $L_1$ with constrained decoding generates queries $\{Q^1, \dots, Q^M\} = \{\mathcal{Q}^1, \dots, \mathcal{Q^M}\}$.
\end{lemma}
}

\begin{proof}
By the contrapositive of Lemma~\ref{lemma:cd-1.1}, $\tilde{p}(Q) \neq 0$ implies $Q \in \mathcal{Q}$. Hence, any query generated under constrained decoding is a valid query. $\{Q^1, \dots, Q^M\} \subseteq \{\mathcal{Q}^1, \dots, \mathcal{Q^M}\}$. Equality follows from cardinality.

\end{proof}

\section{Experimental details} \label{sec:experiemnt_setup}
All experiments are done on a single machine with one 40GB A100 GPU, 85G RAM 12vCPUs. We use Neo4j Community Edition database with default database configurations. Total time for reproducing results for one datasets (e.g STaRK-prime), from loading raw data into the database, finetuning and evaluting metrics, is around 18 hours.

\subsection{Few-shot prompt for entity resolution}
\subsubsection{STaRK-prime}
\begin{lstlisting}
Question : "Which anatomical structures lack the expression of genes or proteins involved in the interaction with the fucose metabolism pathway?" 

Answer : "fucose metabolism"

Question : "What liquid drugs target the A2M gene/protein and bind to the PDGFR-beta receptor?"

Answer : "A2M gene/protein|PDGFR-beta receptor"

Question : "Which genes or proteins are linked to melanoma and also interact with TNFSF8?"

Answer : "melanoma|TNFSF8"

LLM insturction: "You are a knowledgeable assistant which identifies medical entities in the given sentences. Separate entities using '|'."
\end{lstlisting}

\subsubsection{STaRK-mag}
\begin{lstlisting}
Question : "Could you find research articles on the measurement of radioactive gallium isotopes disintegration rates?"

Answer : "FieldOfStudy:measurement of radioactive gallium isotopes disintegration rates"

Question : "What research on water absorption in different frequency ranges have been referenced or deemed significant in the paper entitled 'High-resolution terahertz atmospheric water vapor continuum measurements'"

Answer: "FieldOfStudy:water absorption in different frequency ranges
Paper:High-resolution terahertz atmospheric water vapor continuum measurements"

Question : "Publications by Point Park University authors on stellar populations in tidal tails"

Answer : "Institution:Point Park University\nFieldOfStudy:stellar populations in tidal tails"

Question : "Show me publications by A.J. Turvey on the topic of supersymmetry particle searches."

Answer: "Author:A.J. Turvey\nField of study: supersymmetry particle searches"

LLM instruction: "You are a smart assistant which identifies entities in a given questions. There are institutions, authors, fields of study and papers."
\end{lstlisting}

\subsection{K-hop query path templates}
For both STaRK-prime and STaRK-mag, we use three path templates which are 1-hop, 2-hop and length two paths that connect two entities to curate training Cypher queries. Figure~\ref{fig:cypher-example} shows the query for 2-hop. For 1-hop we substitue the pattern matching with:
\begin{lstlisting}[language=cypher]
MATCH (src {name: srcName})-[r]-(tgt)
\end{lstlisting}
and for length-two paths connecting entities, our query is:
\begin{lstlisting}[language=cypher]
UNWIND $src_names AS srcName1
UNWIND $src_names AS srcName2
MATCH (src1 {name: srcName1})-[r1]-(tgt)-[r2]-(src2 {name: srcName2}) 
WHERE src1 <> src2
RETURN labels(src1)[0] AS label1, src1.name AS name1, type(r1) AS type1, labels(tgt)[0] AS label2, type(r2) AS type2, labels(src2)[0] AS label3, src2.name AS name3, count(DISTINCT tgt) AS totalCnt
\end{lstlisting}

\subsection{Finetuning experimental setups}
\subsubsection{Training LLM $L_1$}
For finetuning LLM $L_1$ we use huggingface transformer library with LoRA.
For base model neo4j/text2cypher-gemma-2-9b-it-finetuned-2024v1 (license: apache-2.0)  and  google/gemma-2-9b-it (license: gemma), we have 
\begin{lstlisting}[language=python]
start_of_generation_tokens = "<start_of_turn>assistant\n"
end_of_generation_token = "<eos>"
bnb_config = BitsAndBytesConfig(load_in_4bit=True, bnb_4bit_use_double_quant=True, bnb_4bit_quant_type="nf4", 
bnb_4bit_compute_dtype=torch.bfloat16, )
tokenizer.padding_side = "right"
model = AutoModelForCausalLM.from_pretrained(
                model_dir,
                quantization_config=bnb_config,
                torch_dtype=torch.bfloat16,
                attn_implementation="eager",
                low_cpu_mem_usage=True,
            )
lora_config = LoraConfig(r=64, lora_alpha=64, target_modules=None, lora_dropout=0.05, bias="none", task_type="CAUSAL_LM", )
\end{lstlisting}
For base model meta-llama/Llama-3.1-8B-Instruct (license: llama3.1) used in ablation, the differences are in the tokenization and padding:
\begin{lstlisting}[language=python]
start_of_generation_tokens = "<|start_header_id|>model<|end_header_id|>\n"
end_of_generation_token = "<|eot_id|>"
tokenizer.pad_token = '<|finetune_right_pad_id|>'
tokenizer.padding_side = "right"
\end{lstlisting}

The supervised finetuning training loop is configured as:
\begin{lstlisting}[language=python]
 sft_config = SFTConfig(auto_find_batch_size=True,
                       gradient_accumulation_steps=1, #8
                       dataset_num_proc=8,
                       num_train_epochs=1,
                       learning_rate=2e-5,
                       optim="paged_adamw_8bit",
                       max_seq_length=max_seq_len,
                       eval_strategy="epoch",
                       save_strategy="epoch",
                       logging_steps=10,
                       output_dir=model_save_dir,
                       load_best_model_at_end=True,
                       )
\end{lstlisting}

\subsubsection{Training LLM $L_2$}
For finetuning LLM $L_2$ we use unsloth\cite{unsloth}. For the base model meta-llama/Llama-3.1-8B-Instruct (license: llama3.1), our configurations are:
\begin{lstlisting}[language=python]
INSTRUCTION_TEMPLATE = "<|start_header_id|>user<|end_header_id|>\n"
RESPONSE_TEMPLATE = "<|start_header_id|>model<|end_header_id|>\n"
EOS = "<|eot_id|>"
RIGHT_PAD_TOKEN = '<|finetune_right_pad_id|>'
ANSWER_SEPARATOR = '|'
PRIME_MAX_SEQUENCE_LENGTH = 15_000
MAG_MAX_SEQUENCE_LENGTH = 15_000
MAX_NEW_TOKENS = 100
INSTRUCTION = ("Given the information below, return the correct nodes for the following question: {question}\n Retrieved information:\n{info}\n")

model, tokenizer = FastLanguageModel.from_pretrained(
            model_name=model_dir,
            max_seq_length=max_sequence_length,  # None
            dtype=torch.bfloat16,
            load_in_4bit=True,
        )
tokenizer.padding = True
tokenizer.pad_token = RIGHT_PAD_TOKEN
tokenizer.padding_side = 'right'

model = FastLanguageModel.get_peft_model(
                model=model,
                r=64,
                target_modules=["q_proj", "k_proj", "v_proj", "o_proj", "gate_proj", "up_proj", "down_proj", ],
                lora_alpha=64,
                bias="none",
            )

sft_config = SFTConfig(
            per_device_train_batch_size=1,
            per_device_eval_batch_size=1,
            # auto_find_batch_size=True,
            dataset_num_proc=8,
            bf16=True,
            num_train_epochs=1,
            gradient_accumulation_steps=4,
            warmup_steps=5,
            learning_rate=2e-5,
            optim="adamw_8bit",
            weight_decay=0.01,
            lr_scheduler_type="linear",
            logging_steps=1,
            eval_strategy="epoch",
            save_strategy="epoch",
            output_dir=model_save_dir,
            load_best_model_at_end=True,
        )

\end{lstlisting}

\section{Analysis of alternative query plans} \label{ap:query-plan}
An alternative valid but suboptimal query plan for the same Cypher query in Table~\ref{tab:query-planner} is shown in Table~\ref{tab:bad-query-planner}. The query optimiser rules out executing this less optimal plan. Adhoc retrieval methods in GraphRAG that do not use the underlying databases either require essentially a rewrite of the query engine in-memory, or risks having suboptimal retrievals.
\begin{table*}
    \centering
    \caption{A valid but suboptimal query plan for an example retrieval query: \lstinline|MATCH (x1:GeneOrProtein {name: "CYP3A4"})-[r: ENZYME]-(x2: Drug)-[r2: INDICATION]-(x3: Disease {name: "strongyloidiasis"}) RETURN x2.name|. Compared with the optimal plan in Table~\ref{tab:query-planner}, this plan has both more operators and more costly ones.}
    \label{tab:bad-query-planner}
    \ttfamily 
    \resizebox{\textwidth}{!}{ 
    \begin{tabular}{|l|l|l|l|}
        \toprule
        Operator    & Id  & Details     & Estimated Rows\\
        & & & \\
        \midrule
        +ProduceResults  & 0   & n`x2.name`    & 63     \\
        \cmidrule(lr){2-4}
        +Projection      & 1   & x2.name AS `x2.name`    & 63  \\
        \cmidrule(lr){2-4}
        |+NodeHashJoin   & 2   & x2   & 109441  \\
        \cmidrule(lr){2-4}
        ||+Filter          & 3   & x1.name = \$autostring\_0 AND x1:GeneOrProtein   & 51795  \\
        \cmidrule(lr){2-4}
        ||+Expand(All)     & 4   & (x2)-[r2:ENZYME]-(x1)      & 19893   \\
        \cmidrule(lr){2-4}
        ||+NodeByLabelScan & 5   & x2:Drug      & 7957  \\
        \cmidrule(lr){2-4}
        |+Filter          & 6   & x2:Drug  & 7957           \\
        \cmidrule(lr){2-4}
        |+Expand(All)     & 7   & (x3)-[r:INDICATION]-(x2)       & 52521    \\
        \cmidrule(lr){2-4}
        |+Filter          & 8   & x3.name = \$autostring\_1  & 51240        \\
        \cmidrule(lr){2-4}
        +NodeByLabelScan  & 9   & x3:Disease               & 17080           \\
        \bottomrule
    \end{tabular}
    }
\end{table*}

\section{Ablation of choices of LLMs} \label{ap:ablation}
\begin{figure}
    \centering
    \begin{subfigure}{0.8\textwidth}
        \centering
        \includegraphics[width=\linewidth]{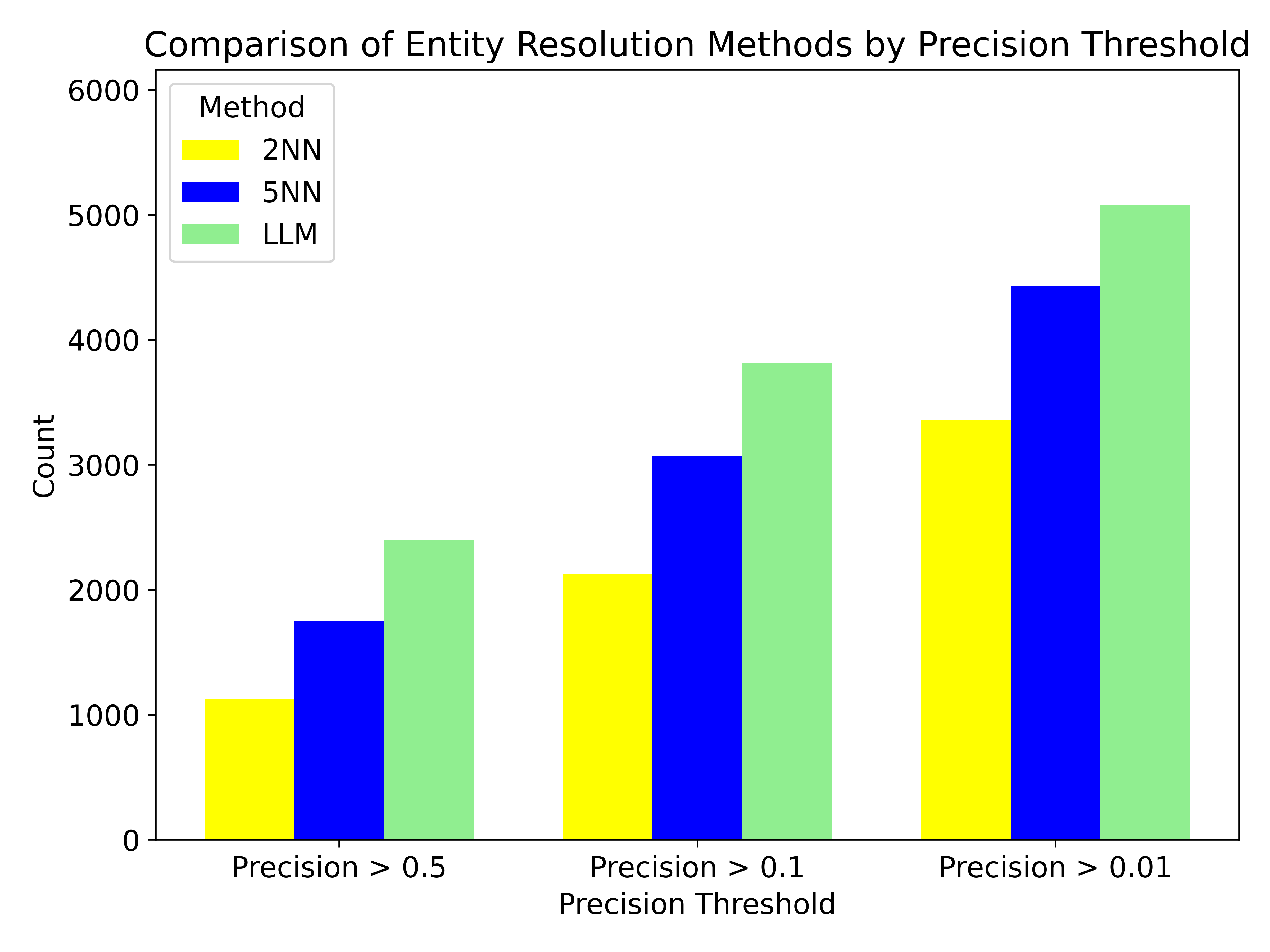}
        \caption{The number of questions that map to good Cyphers when using different entity resolution methods.}
    \end{subfigure}
    \caption{The impact of entity resolution on the quality of Cypher queries.}
    \label{fig:entity-resolution-ablation}
\end{figure}

Given $\{Q_i, A_i\}$ as input data, we prepare a set of $\{Q_i, C_i\}$ as a set of training data to finetune the LLM to generate optimal Cypher queries. We first few-shot prompt an LLM to identify entities and then reconcile the identified names against the DB. Next, by using the graph schema, we obtain the set of all possible Cypher queries with type predicates k-hop around or connecting the entity nodes. The quality of the identified entities is therefore implicitly measured by the best Cypher that it allows. Figure~\ref{fig:entity-resolution-ablation} shows that using an LLM for entity resolution does produce better entities that lead to more high-recall ground-truth Cypher queries with more accurate subgraphs returned.

We also finetune different base LLMs for $L_1$ as shown in Table~\ref{tab:ablation-l1}.
\begin{table}
	\centering
	\caption{Metircs for STaRK-prime when using $L_1$ only. Two different base models are finetuned. Both finetuned models achieve beyond SOTA results on every metric.}
	\label{tab:ablation-l1}
	\begin{tabular}{c|cccc}
		\toprule
            & \multicolumn{4}{c}{STARK-PRIME}  \\
		  & Hit@1 & Hit@5 & R@20 & MRR  \\
            Llama3.1-8b-Instruct & 43.88 & 64.44 & 70.60 & 52.85  \\
            Gemma2-9b-it & 48.27 & 68.23 & 74.46 & 57.12  \\
            \bottomrule
	\end{tabular}
\end{table}

\end{document}